\documentclass[lettersize,journal]{IEEEtran}
\usepackage{amsmath,amsfonts}
\usepackage{algorithmic}
\usepackage{algorithm}
\usepackage{array}
\usepackage[caption=false,font=normalsize,labelfont=sf,textfont=sf]{subfig}
\usepackage{textcomp}
\usepackage{stfloats}
\usepackage{url}
\usepackage{verbatim}
\usepackage{graphicx}
\usepackage{cite}
\usepackage[hidelinks]{hyperref}
\usepackage[table]{xcolor}
\definecolor{softgreen}{RGB}{232,245,233}

\usepackage[table]{xcolor}   

\definecolor{mygreen}{RGB}{170, 230, 180}

\usepackage{booktabs}   
\usepackage{tabularx}   
\usepackage{array}      
\usepackage{ragged2e}   
\usepackage[hidelinks]{hyperref} 

\usepackage{booktabs}        
\usepackage{multirow}        
\usepackage{threeparttable}  

\begin{document}

\title{Chinese Labor Law Large Language Model Benchmark}


\author{Zixun Lan, Maochun Xu, Yifan Ren, Rui Wu, Jianghui Zhou, Xueyang Cheng, Jian'an Ding, Xinheng Wang, Mingmin Chi, Fei Ma
\thanks{Zixun Lan and Maochun Xu contributed equally to this work.}%
\thanks{Zixun Lan and Xinheng Wang are with Department of Mechatronics and Robotics, Xi’an Jiaotong–Liverpool University, Suzhou, China.
}
\thanks{Maochun Xu, Yifan Ren and Fei Ma are with Department of Applied Mathematics, Xi’an Jiaotong–Liverpool University, Suzhou, China.}
\thanks{Rui Wu is with Hansheng Lawyers Building.}
\thanks{Jianghui Zhou is with Suzhou Gewu Digital Technology Co., Ltd.}
\thanks{Xueyang Cheng and Jian’an Ding are with Kenneth Wang School of Law, Soochow University, Suzhou, China.}
\thanks{Mingmin Chi is with College of Computer Science and Artificial Intelligence, Fudan University, Shanghai, China.}
\thanks{Fei Ma is the Corresponding author.}
}



\maketitle

\begin{abstract}
Recent advancements in large language models (LLMs) have led to substantial progress in domain-specific applications, particularly within the legal domain. Although general-purpose models such as GPT-4 exhibit promising performance on basic legal tasks, they often struggle with specialized subdomains that demand precise legal knowledge, complex reasoning, and contextual sensitivity. To address these limitations, this paper presents \textbf{LaborLawLLM}, a legal large language model specifically tailored to the labor law domain—a subfield marked by intricate statutory structures, frequent disputes, and high real-world impact. We construct \textit{LabourLawBench}, a comprehensive benchmark comprising diverse labor law tasks such as legal provision citation, knowledge-based question answering, case classification, compensation computation, named entity recognition, and legal case analysis. Our evaluation framework integrates both objective metrics (e.g., ROUGE-L, accuracy, F1, soft-F1) and subjective assessments based on GPT-4 scoring. Experimental results demonstrate that \textbf{LaborLawLLM} significantly outperforms both general-purpose and existing legal-specific LLMs across all task categories. This work not only fills a key research gap in labor law-specific legal AI but also offers a scalable methodology for developing specialized LLMs in other legal subfields, thereby enhancing the accuracy, reliability, and societal value of legal AI applications. 
\end{abstract}

\begin{IEEEkeywords}
Chinese labor law, legal natural language processing, large language models, domain-specific fine-tuning, benchmark dataset, statute recall, legal reasoning, case analysis
\end{IEEEkeywords}

\section{Introduction}
In recent years, with the rapid development of artificial intelligence technology, large language models (LLMs) have made significant breakthroughs in the field of natural language processing and have been successfully applied in various professional fields such as healthcare, finance, and education \cite{yao2024lawyer}. The legal field, due to its high degree of specialization, logical rigor, and practical significance, has gradually become an important direction for the application and research of LLMs \cite{yue2024lawllm, zhou2024lawgpt}. Specifically, LLMs can efficiently process vast amounts of legal texts and case law, automate the generation of legal documents, and conduct precise legal research and analysis, significantly improving the efficiency and accuracy of legal services. By optimizing the processes of legal document drafting, case analysis, and contract review, LLMs demonstrate immense potential in both litigation and non-litigation tasks, becoming a key driver in the digital transformation of the legal industry.

\begin{figure*}[!t]
  \centering
    \caption{Performance of baselines across tasks.}
  \label{fig:rader_renwu}
  \includegraphics[width=0.9\linewidth]
  {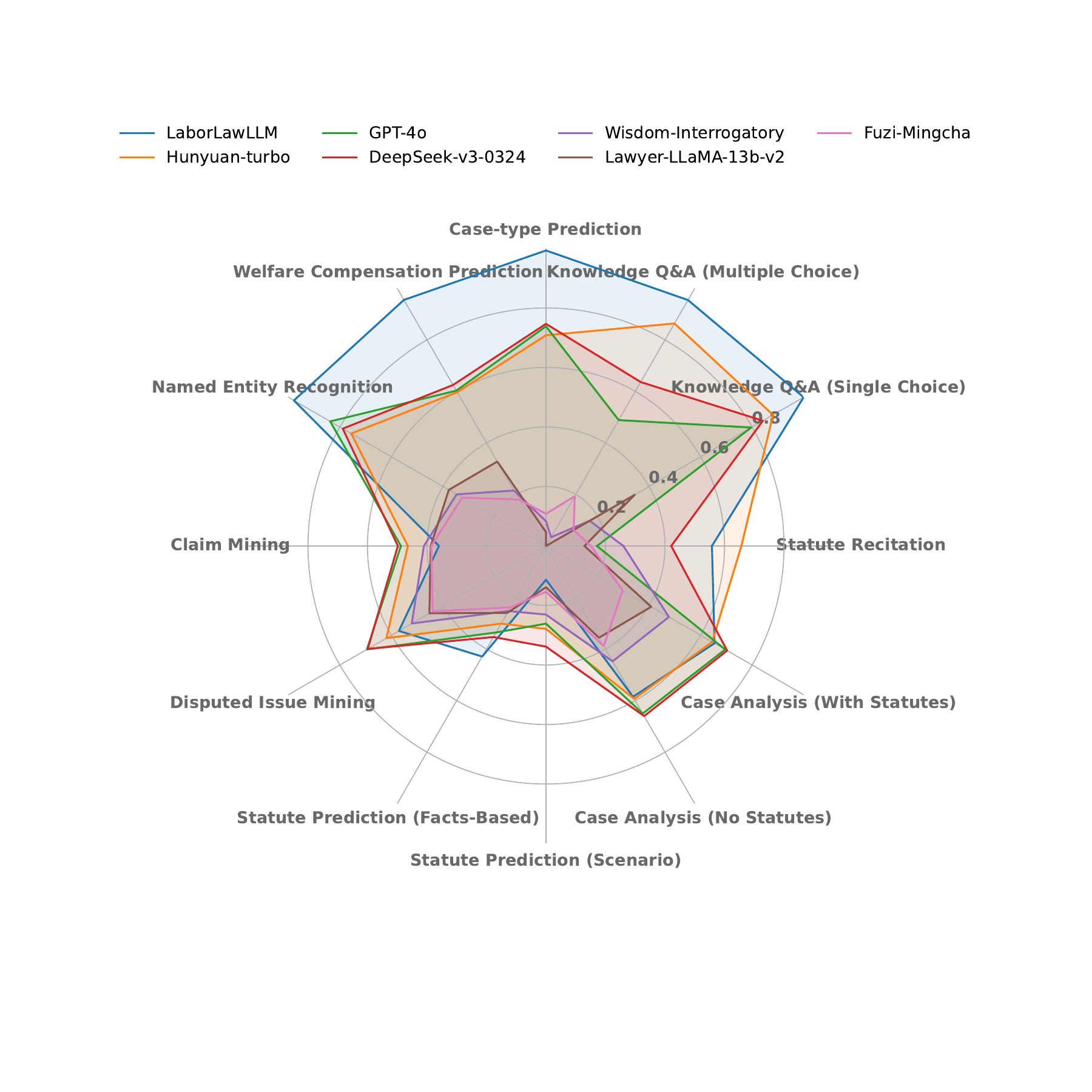}
  \vspace*{-3cm}
\end{figure*}

\begin{figure*}[!t]
  \centering
  \caption{Performance of baselines across case types.}
  \label{fig:rader_case}
  \includegraphics[width=0.9\linewidth]{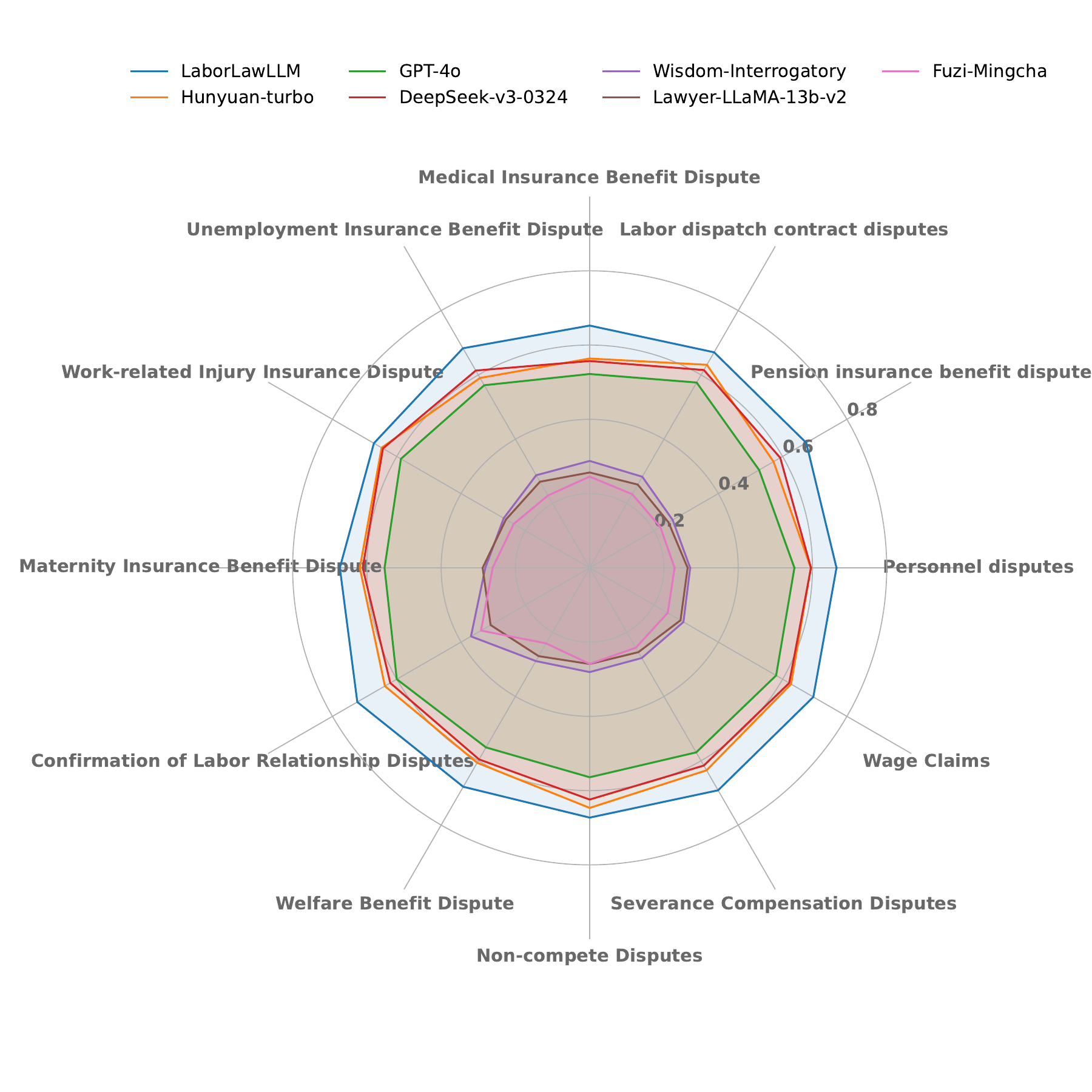}
  \vspace*{-2cm}
\end{figure*}

A variety of legal-domain large language models (LLMs) have been proposed, including ChatLaw \cite{cui2023chatlaw}, DISC-LawLLM \cite{yue2023disc}, LaWGPT \cite{song2023lawgpt, zhou2024lawgpt} and Lawyer-LLaMA \cite{huang2023lawyer}. These models utilise large-scale legal corpora and domain-specific fine-tuning strategies to substantially enhance legal knowledge representation and reasoning capabilities. For example, LaWGPT, built upon Chinese-LLaMA-7B, outperforms general-purpose models across a range of Chinese legal tasks \cite{zhou2024lawgpt}. DISC-LawLLM integrates a legal reasoning fine-tuning dataset with an external knowledge retrieval module, significantly improving robustness in practical legal scenarios  \cite{yue2023disc}. ChatLaw, based on Ziya-LLaMA-13B, incorporates a Mixture of Experts (MoE) architecture and a knowledge graph-enhanced multi-agent collaboration framework, surpassing GPT-4 in certain legal consultation tasks \cite{cui2023chatlaw}. Lawyer-LLaMA adopts continual pre-training on legal texts and expert-guided instruction tuning to effectively mitigate hallucination in legal text generation \cite{huang2023lawyer}. Additionally, models such as HanFei \cite{wen2023hanfei}, wisdomInterrogatory \cite{zhihai2023wisdominterrogatory}, Fuzi-Mingcha \cite{wu2023fuzi}, and LexiLaw \cite{li2023lexilaw} are tailored for specific applications such as legal question answering and judicial decision analysis. These legal LLMs typically employ parameter scales ranging from 7B to 13B, aligning with general-purpose models to balance task performance and computational efficiency. These representative models are later compared in our benchmark. 
As illustrated in Figure~\ref{fig:rader_renwu} and Figure~\ref{fig:rader_case}, 
we evaluate multiple baselines across twelve labor‐law tasks and twelve case types, 
revealing notable performance gaps among general‐purpose and domain‐specific legal LLMs.

Although existing legal-domain large language models (LLMs) have demonstrated promising performance on general legal tasks such as legal consultation, case classification, and statute retrieval \cite{zhou2024lawgpt, yue2024lawllm, cui2023chatlaw}, most current research remains focused on broad, undifferentiated legal scenarios. This overlooks the substantial heterogeneity that exists across subfields within the legal domain. Branches such as labor law, intellectual property law, tax law, and criminal law are characterized by highly specialized terminologies, normative logic, and distinct practical requirements. Due to this complexity and domain-specificity, general-purpose LLMs often struggle to capture the embedded reasoning structures and specialized knowledge systems required for accurate legal analysis in these subdomains. Studies have shown that such models are prone to generating logically inconsistent or legally inaccurate outputs—so-called hallucinations—when applied to fine-grained, context-dependent legal tasks, which poses risks of misinformation in real-world legal applications \cite{huang2023lawyer, du2021glm}. This issue stems in part from the insufficient coverage of subdomain-specific statutes, precedents, and practical norms in general training corpora \cite{bommarito2021lexnlp}, and in part from the difficulty of modeling the implicit assumptions and hierarchical constraints embedded in legal reasoning (e.g., syllogistic logic, statutory interpretation) using standard language modeling techniques \cite{chalkidis2020legal}. As a result, developing more specialized LLMs tailored to individual legal subfields—supported by domain-specific corpora and supervised fine-tuning aligned with subfield practices—has become a critical direction for advancing the 
applicability and reliability of AI in legal contexts.

This study proposes the development of a large language model specifically designed for the domain of labor law, with the aim of enhancing the applicability of legal AI in highly specialized scenarios. As a legal subfield closely connected to social welfare and workers' rights, labor law carries significant practical importance and is associated with a large volume of litigation and dispute resolution cases. The legal framework of labor law—including labor contract law, arbitration regulations, administrative rules, and judicial interpretations—is complex and highly specialized \cite{davidov2006boundaries}. Compared with other legal domains, labor law cases often involve the precise application of legal provisions, detailed calculations of compensation, and a deep understanding of real-world labor situations, all of which place high demands on a model's legal knowledge coverage, logical reasoning ability, and fact-law alignment . However, existing general-purpose and legal-domain language models often fall short in addressing labor law tasks, exhibiting problems such as inaccurate citation of statutes, insufficient reasoning rigor, and lack of professional reliability in their outputs, which significantly limits their practical value in this field \cite{fei2023lawbench}. Developing a labor law-specific LLM is therefore not only a necessary 
step to fill the gap in current legal AI
applications, but also a promising direction for advancing legal AI toward higher levels of specialization and real-world utility.

This study addresses the limitations of general-purpose and legal-domain language models in specialized areas of law by proposing and implementing a systematic approach for building a domain-specific large language model (LLM) for labor law. The main contributions of this work are as follows:

\begin{itemize}
  \item We conduct a comprehensive analysis of the shortcomings of existing general and legal LLMs in handling specialized legal tasks. Based on this, we identify the necessity of developing subdomain-specific legal models and construct a high-quality training dataset covering core labor law tasks. A dedicated pre-training and fine-tuning strategy is employed to develop a labor law–specific LLM.
  
  \item We design a task framework and evaluation protocol tailored to the labor law domain, including statutory citation, dispute identification, and compensation calculation. This includes a diverse set of objective and subjective evaluation metrics. To the best of our knowledge, this is the first systematic evaluation of a legal LLM in the labor law subfield, filling an important gap in existing research.
  
  \item Through rigorous experiments, we demonstrate that the proposed labor law LLM significantly outperforms both general-purpose and general legal LLMs on multiple labor-specific tasks. The results validate the effectiveness and practical value of subdomain-specific modeling and provide insights and methodologies that are transferable to other areas of legal AI development.
\end{itemize}

In summary, this research not only contributes to a deeper theoretical understanding of how large language models adapt to domain-specific legal knowledge and reasoning, but also offers a practical, scalable solution for improving the quality of legal services, promoting equitable access to legal resources, and strengthening the protection of workers' rights. The framework and findings presented here carry substantial implications for both academic research and real-world legal practice.

\section{Related Work}

\subsection{Chinese Legal Domain-Specific Language Models}

In recent years, Chinese legal domain-specific large language models (LLMs) have achieved notable advances in tasks such as legal text generation, consultation, and judgment prediction. Broadly speaking, existing approaches to developing legal LLMs can be grouped into two categories. The first focuses on fine-tuning general-purpose pretrained models for legal applications \cite{yue2023disc, zhou2024lawgpt, li2023lexilaw, chen2023llmworkchina}. The second emphasizes (continued) pretraining on large-scale legal corpora to enhance models' understanding of legal language and reasoning patterns \cite{wu2023fuzi, dai2023laiw}.

The first category includes models such as DISC-LawLLM \cite{yue2023disc}, LawGPT\_zh \cite{zhou2024lawgpt}, LexiLaw \cite{li2023lexilaw}, ChatLaw \cite{cui2023chatlaw}, and Lawyer-LLaMA \cite{huang2023lawyer}. These models are typically built upon base models like ChatGLM~\cite{chatglm2024}, LLaMA~\cite{cui2023chatlaw}, or Baichuan~\cite{baichuan2023}, and are fine-tuned using structured legal data, such as question–answer pairs, statutes, and court decisions. To address the limitation of incomplete legal knowledge during pretraining, some models incorporate external legal knowledge bases or retrieval-augmented generation (RAG) modules to improve accuracy and relevance \cite{yue2023disc, ma2025sddlawllm, park2025lrage}. These methods are efficient and easy to deploy, but their performance is highly dependent on data quality and instruction format. Moreover, they often struggle with complex legal reasoning tasks that require multi-step inference.

The second category focuses on enhancing legal knowledge representation through legal-domain pretraining. Representative models include LaWGPT \cite{song2023lawgpt}, WisdomInterrogatory \cite{zhihai2023wisdominterrogatory}, and HanFei \cite{wen2023hanfei}. These models leverage large volumes of legal texts, including statutes, court rulings, and case summaries, to improve performance in tasks, such as statute prediction, case classification, and element extraction. By aligning model parameters more closely with the linguistic and semantic patterns of legal discourse, domain pretraining tends to yield stronger generalization across unseen legal tasks compared to direct fine-tuning. However, despite these advantages, such models still focus primarily on text generation and lack explicit mechanisms for rigorous legal reasoning, structured causal analysis, and normative or ethical judgment. These limitations narrow their applicability in complex legal domains, particularly those demanding multi-step deductive reasoning, transparent justificatory frameworks, and alignment with established jurisprudential principles.

Overall, the current focus of Chinese legal LLMs remains on legal knowledge recall and text-based question answering (e.g., LexiLaw \cite{li2023lexilaw}, ChatLaw \cite{cui2023chatlaw}, LaWGPT \cite{song2023lawgpt}, HanFei \cite{wen2023hanfei}). In contrast, relatively little attention has been devoted to capturing the formal structure of legal reasoning and causal logic \cite{du2021glm, cui2023chatlaw, wen2023hanfei}. Yet, for real-world legal applications—especially those involving complex disputes—it is crucial that models demonstrate coherent reasoning and the ability to perform multi-step, hierarchical decision-making, capabilities that remain insufficiently developed in current approaches.

\subsection{Evaluation Benchmarks for Chinese Legal LLMs}

To systematically evaluate legal LLMs across different legal tasks and cognitive levels, several evaluation benchmarks have emerged in recent years. LAiW \cite{dai2023laiw} introduces a Chinese legal benchmark that incorporates layered legal reasoning depth, dividing model capabilities into three levels: Basic Information Retrieval (BIR), Legal Fundamental Inference (LFI), and Complex Legal Application (CLA). This framework underscores persistent limitations in constructing coherent reasoning chains. LawBench \cite{fei2023lawbench} encompasses 20 tasks within the Chinese legal system and evaluates legal knowledge recall, comprehension, and application across classification, regression, extraction, and generation tasks. LexEval \cite{li2024lexeval}, one of the most comprehensive benchmarks to date, comprises 23 tasks and 14,150 questions, and places particular emphasis on ethical judgment and logical consistency, making it an important reference for the evaluation of legal LLMs. In addition, DISC-Law-Eval \cite{yue2023disc} proposes a hybrid evaluation approach that combines objective metrics with human evaluation to assess factual accuracy and logical soundness. SimuCourt \cite{he2024agentscourt} simulates court debates and legal reasoning processes to evaluate the model's ability to perform judicial analysis and decision-making, bringing evaluation closer to real legal practice.

Despite these advancements, most existing benchmarks continue to emphasize legal knowledge recall and answer correctness, while providing relatively limited evaluation of models' abilities in complex legal reasoning, causality analysis, ethical assessment, and judgment explanation. This gap is especially evident in domains such as labor law, where disputes frequently involve multiple parties and nuanced statutory interpretation. While frameworks like LAiW and LawBench offer solid assessments of foundational legal knowledge, they remain limited in evaluating reasoning depth and practical judicial applicability \cite{hyde2006labor}. As a result, legal professionals have expressed growing concerns about the reliability of LLMs in high-stakes legal contexts.

\subsection{Motivation: Toward a Labor Law Specific Evaluation Framework}

To address these gaps, our work proposes a benchmark specifically tailored to the labor law domain, which involves unique dispute types, legal structures, and ethical dilemmas. We design a suite of 12 evaluation tasks covering statute recall, legal knowledge Q\&A, cause of action prediction, dispute focus extraction, and legal reasoning analysis. This benchmark fills the current gap in legal evaluation for labor law and emphasizes the model's ability to identify legal elements, construct reasoning chains, and reach reasonable legal conclusions in realistic labor dispute scenarios. As such, it serves as an important extension to existing Chinese legal LLM evaluation frameworks and lays the foundation for developing labor law LLMs with strong reasoning and real-world applicability.

\section{Data}

\begin{table*}[!t]
\centering
\caption{Distribution of the benchmark dataset by task (T$_1$--T$_{12}$) and case type (C$_1$--C$_{12}$). Symbol ``---'' indicates tasks not divided by case type.}
\label{tab:benchmark_distribution}

\setlength{\tabcolsep}{3pt}          
\renewcommand{\arraystretch}{1.05}   

\resizebox{\linewidth}{!}{%
\begin{tabular}{l*{12}{c}c}
\hline
\textbf{Task} & C$_1$ & C$_2$ & C$_3$ & C$_4$ & C$_5$ & C$_6$ & C$_7$ & C$_8$ & C$_9$ & C$_{10}$ & C$_{11}$ & C$_{12}$ & \textbf{Total} \\
\hline
T$_1$ Statute Recitation               & \multicolumn{12}{c}{---} & 3165 \\
T$_2$ Knowledge QA (Single Choice)     & \multicolumn{12}{c}{---} & 177  \\
T$_3$ Knowledge QA (Multiple Choice)   & \multicolumn{12}{c}{---} & 88   \\
T$_4$ Case-Type Prediction             & 50 & 50 & 50 & 50 & 50 & 50 & 50 & 50 & 50 & 50 & 50 & 50 & 600 \\
T$_5$ Welfare Compensation Prediction  & 50 & 50 & 50 & 50 & 50 & 50 & 50 & 50 & 50 & 50 & 50 & 50 & 600 \\
T$_6$ Named Entity Recognition         & 50 & 50 & 50 & 50 & 50 & 50 & 50 & 50 & 50 & 50 & 50 & 50 & 600 \\
T$_7$ Claim Mining                     & 50 & 50 & 50 & 50 & 50 & 50 & 50 & 50 & 50 & 50 & 50 & 50 & 600 \\
T$_8$ Disputed Issue Mining            & 50 & 50 & 50 & 50 & 50 & 50 & 50 & 50 & 50 & 50 & 50 & 50 & 600 \\
T$_9$ Statute Prediction (Facts-Based) & 50 & 50 & 50 & 50 & 50 & 50 & 50 & 50 & 50 & 50 & 50 & 50 & 600 \\
T$_{10}$ Statute Prediction (Scenario) & \multicolumn{12}{c}{---} & 670  \\
T$_{11}$ Case Analysis (No Statutes)   & 32 & 32 & 32 & 32 & 32 & 32 & 32 & 32 & 32 & 32 & 32 & 32 & 384 \\
T$_{12}$ Case Analysis (With Statutes) & 32 & 32 & 32 & 32 & 32 & 32 & 32 & 32 & 32 & 32 & 32 & 32 & 384 \\
\hline
\end{tabular}
}
\end{table*}


\begin{table*}[!t]
\centering
\caption{Name mapping of the 12 case types (C$_1$--C$_{12}$).}
\label{tab:c1c12_vertical}

\resizebox{\linewidth}{!}{%
\begin{tabularx}{\linewidth}{@{}l X l X@{}}
\toprule
\textbf{Code} & \textbf{Case Type} & \textbf{Code} & \textbf{Case Type} \\
\midrule
C1  & Personnel disputes                      & C7  & Maternity insurance benefit disputes \\
C2  & Pension insurance benefit disputes      & C8  & Labor relationship confirmation disputes \\
C3  & Labor dispatch contract disputes        & C9  & Welfare benefit disputes \\
C4  & Medical insurance benefit disputes      & C10 & Non-compete disputes \\
C5  & Unemployment insurance benefit disputes & C11 & Severance compensation disputes \\
C6  & Work-injury insurance benefit disputes  & C12 & Wage claims \\
\bottomrule
\end{tabularx}%
}
\end{table*}

\begin{table*}[!t]
\centering
\caption{Overview of Tasks in the Labor Law Evaluation Benchmark.}
\label{tab:legal_tasks}
\renewcommand{\arraystretch}{1.2}
\setlength{\tabcolsep}{10pt}
\resizebox{\textwidth}{!}{
\begin{tabular}{l l l l} 
\hline
\textbf{Task} & \textbf{Task Type} & \textbf{Metric} & \textbf{Data Size} \\
\hline
T1 Legal Provision Memorization & Generation & ROUGE-L & 3165 \\
T2 Knowledge Q\&A (Single Choice) & Single-label Classification & Accuracy & 620 \\
T3 Knowledge Q\&A (Multiple Choice) & Multi-label Classification & Accuracy & 310 \\
T4 Cause of Action Prediction & Single-label Classification & Accuracy & 600 \\
T5 Welfare Compensation Prediction & Multi-label Classification & F1 & 600 \\
T6 Named Entity Recognition & Extraction & Soft-F1 & 600 \\
T7 Claim Mining & Extraction & GPT-O1 & 600 \\
T8 Dispute Focus Extraction & Extraction & GPT-O1 & 600 \\
T9 Legal Provision Prediction (Fact-based) & Regression & GPT-O1 & 600 \\
T10 Legal Provision Prediction (Scenario-based) & Regression & ROUGE-L & 670 \\
T11 Case Analysis (w/o Legal Provisions) & Generation & GPT-O1 & 384 \\
T12 Case Analysis (w/ Legal Provisions) & Generation & GPT-O1 & 384 \\
\hline
\end{tabular}
}
\end{table*}

We build a balanced benchmark dataset for the labour-law domain covering twelve tasks (T$1$–T${12}$), each spanning twelve labour-law sub-causes (C$1$–C${12}$).
The dataset distribution by task and case type is summarized in Table~\ref{tab:benchmark_distribution}, with corresponding case-type names listed in Table~\ref{tab:c1c12_vertical}, and an overview of task types and evaluation metrics provided in Table~\ref{tab:legal_tasks}.
For category-dependent tasks (T$4$–T$9$, T${11}$–T${12}$), we construct each task with a fixed quota of 50 cases per category (12×50=600), as shown in Table~\ref{tab:benchmark_distribution}.
For \emph{category-independent} tasks (e.g., statute recitation, scenario-based provision prediction), sizes follow source availability (e.g., T$1$: 3{,}165 provisions; T${10}$: 670 scenarios).
All tasks adopt a unified record format—\textbf{Instruction}, \textbf{Question}, and \textbf{Answer}—with task-specific output rules.
In the subsections that follow, we describe the construction, input–output format, and evaluation protocol for each task, following the balanced design summarized in Tables~\ref{tab:benchmark_distribution}–\ref{tab:legal_tasks}.

\subsection{Statute Recitation Task}

To evaluate verbatim memorization and exact retrieval of statutory text, we construct a statute recitation dataset (corresponding to T1 in Table~\ref{tab:legal_tasks}) derived from Chinese labor law. Specifically, we select 3,165 substantive provisions from the \textit{Labor Law of the People's Republic of China} \cite{PRC-Labour-Law-2018} and related legal instruments. These provisions constitute the source material for the statute recitation task in our benchmark.

Each instance is formatted as a triplet—\textbf{Instruction}, \textbf{Question}, and \textbf{Answer}—following the supervised fine-tuning (SFT) convention for instruction-following models:
\begin{itemize}
    \item \textbf{Instruction:} Directs the model to act as a legal expert in labor and social security law and to output the complete statutory text; paraphrasing or explanation is prohibited.
    \item \textbf{Question:} Specifies the query (typically the law name and article number) to be recalled verbatim.
    \item \textbf{Answer:} Provides the exact official wording of the article; precision in terminology and structure is required.
\end{itemize}

A formatted instance is shown table \ref{tab:statute_example}:
\begin{table}[!t]
\centering
\renewcommand{\arraystretch}{1.2}
\setlength{\arrayrulewidth}{0.5pt}
\caption{Instruction format and an example for the Statute Recitation task (T$_1$).}
\label{tab:statute_example}
\begin{tabular}{>{\columncolor{softgreen}}p{0.98\linewidth}}
\hline
\textbf{INSTRUCTION}: You are a legal expert in the field of labor and social security law. Answer the following question by providing only the text of the relevant statutory provision.\\
\hline
\textbf{QUESTION}: What is the content of Article 16 of the \textit{Law of the People's Republic of China on Mediation and Arbitration of Labor Disputes}?\\
\hline
\textbf{ANSWER}: Where a mediation agreement has been reached with respect to the payment of unpaid labor remuneration, medical expenses for work-related injuries, economic compensation, or damages, and the employer fails to perform within the time limit specified in the agreement, the worker may, on the strength of the written mediation agreement, apply to the people's court for a payment order in accordance with law. The people's court shall issue a payment order in accordance with law.\\
\hline
\end{tabular}
\end{table}

In total, the task contains 3,165 fully processed items. This set is used to assess a model's ability to memorize, recall, and faithfully reproduce statutory provisions.

\subsection{Knowledge Q\&A (Single Choice)}

To assess factual understanding of labor law, we construct a single-choice knowledge question dataset (corresponding to T2 in Table~\ref{tab:legal_tasks}) from national-level examination materials. Specifically, we collect items from the \textit{Labor Law} subject in self-taught higher education examinations (2004–2023), the National Judicial Examination (2008–2017), and the Unified National Legal Professional Qualification Exam (2018–2022). We first filter questions relevant to labor law and then retain those with a single correct option.

Each instance is formatted as a triplet—\textbf{Instruction}, \textbf{Question}, and \textbf{Answer}—following the supervised fine-tuning (SFT) convention for instruction-following models:
\begin{itemize}
    \item \textbf{Instruction:} Sets the role as a legal expert in labor and social security law and requires selecting exactly one correct option from A–D. The output must strictly follow the format \texttt{[Correct Answer]X<eoa>}.
    \item \textbf{Question:} Provides the question stem and all four options (A, B, C, D).
    \item \textbf{Answer:} Gives the gold label in the standardized format \texttt{[Correct Answer]X<eoa>} (e.g., \texttt{[Correct Answer]C<eoa>}).
\end{itemize}

In total, this task contains 177 fully processed questions, used to evaluate a model's ability to select legally correct answers under standardized output constraints. A formatted instance is shown below table \ref{tab_single_Choice}:

\begin{table}[!h]
\centering
\caption{Instruction format and an example for the Knowledge Q\&A (Single Choice) task.}
\label{tab_single_Choice}
\renewcommand{\arraystretch}{1.25}
\setlength{\arrayrulewidth}{0.4pt}
\begin{tabular}{>{\columncolor{softgreen}}p{0.97\linewidth}}
\hline
\textbf{INSTRUCTION}: You are a legal expert in the field of labor and social security law. Using your legal knowledge, choose the single correct answer from A, B, C, and D, and write it strictly between \texttt{[Correct Answer]} and \texttt{<eoa>}. For example: \texttt{[Correct Answer]A<eoa>}. You must follow this format exactly.\\
\hline
\textbf{QUESTION}:
Mr.\ Li previously worked for Company A and was subject to a non-fixed working hours system. In January 2012, because Company A was merged into (acquired by) Company B, Mr.\ Li became an employee of Company B and continued to be subject to the non-fixed working hours system. In December 2012, because Mr.\ Li received the lowest score in the annual performance appraisal, Company B decided to terminate Mr.\ Li's labor relationship pursuant to the ``bottom-ranked elimination'' provision in its performance appraisal policy. In November 2013, Mr.\ Li filed an application for labor dispute arbitration, claiming that after the original labor contract expired in March 2012, Company B had never entered into a new written labor contract with him and therefore should pay double wages each month starting from April; and that the termination was unlawful and his job should be reinstated. With respect to issues related to Mr.\ Li's arbitration application, which of the following options is incorrect? [ ]\\[4pt]
A.\ Because the place of performance of the labor contract is inconsistent with the location of Company B, Mr.\ Li may only apply for arbitration to the labor dispute arbitration commission at the place where the labor contract is performed. \quad\\
B.\ When filing the application, an arbitration application form should be submitted; if there are genuine difficulties, an oral application is also permissible. \quad\\
C.\ Company B bears the burden of proof for its claim regarding the termination of the labor contract. \quad\\
D.\ If the labor dispute arbitration commission fails to decide whether to accept the case within the prescribed time limit, Mr.\ Li may bring a lawsuit in court regarding the labor dispute matter.\\
\hline
\textbf{ANSWER}: \texttt{[Correct Answer]A<eoa>} \\
\hline
\end{tabular}
\end{table}

\subsection{Knowledge Q\&A (Multiple Choice)}

To assess the ability to answer labor-law questions with \emph{multiple} correct options, we construct a multiple-choice knowledge dataset (corresponding to T3 in Table~\ref{tab:legal_tasks}) from national-level examinations. Specifically, we curate items from the \textit{Labor Law} subject in self-taught higher education examinations (2004–2024), the National Judicial Examination (2008–2017), and the Unified National Legal Professional Qualification Exam (2018–2022). We first filter questions relevant to labor law and then retain those with more than one correct option.

Each instance is formatted as a triplet—\textbf{Instruction}, \textbf{Question}, and \textbf{Answer}—following the supervised fine-tuning (SFT) convention for instruction-following models:
\begin{itemize}
    \item \textbf{Instruction:} Sets the role as a legal expert in labor and social security law and requires selecting one or more correct options from A–D. The output must \emph{strictly} follow \texttt{[Correct Answer]XY\ldots<eoa>}, where letters are uppercase, deduplicated, and sorted alphabetically without spaces (e.g., \texttt{[Correct Answer]BD<eoa>}).
    \item \textbf{Question:} Provides the question stem and four options (A, B, C, D).
    \item \textbf{Answer:} Gives the gold set of options in the standardized format \texttt{[Correct Answer]XY\ldots<eoa>} (e.g., \texttt{[Correct Answer]BCD<eoa>}).
\end{itemize}

An formatted instance is shown below table \ref{table_Multiple}:


\begin{table}[!h]
\centering
\caption{Instruction format and an example for the Knowledge Q\&A (Multiple Choice) task.}
\label{table_Multiple}
\renewcommand{\arraystretch}{1.25}
\setlength{\arrayrulewidth}{0.4pt}
\begin{tabular}{>{\columncolor{softgreen}}p{0.97\linewidth}}
\hline
\textbf{INSTRUCTION}: You are a legal expert in the field of labor and social security law. This is a multiple-choice question. Using your legal knowledge, choose one or more correct answers from A, B, C, and D, and write them strictly between \texttt{[Correct Answer]} and \texttt{<eoa>}. For example: \texttt{[Correct Answer]AB<eoa>}. You must follow this format exactly.\\
\hline
\textbf{QUESTION}: Although the development of labor legislation differs across countries, their common development trend is [ ] \\[4pt]
A.\ The scope of application of labor law has narrowed. \\ 
B.\ Labor law has become a complete and systematic legal system. \\ 
C.\ The development of international labor legislation has influenced national labor laws.\\ 
D.\ Most countries generally stipulate minimum standards for the main working conditions. \\
\hline
\textbf{ANSWER}: \texttt{[Correct Answer]BCD<eoa>} \\
\hline
\end{tabular}
\end{table}

In total, this task contains 88 fully processed questions and is used to evaluate whether a model can identify \emph{all} legally correct options under standardized output constraints.

\subsection{Case-type Prediction}

To assess single-label classification of legal causes in labor and social security disputes, we construct a cause prediction dataset (corresponding to T4 in Table~\ref{tab:legal_tasks}) from judicial documents publicly released in Jiangsu Province over the past five years. Each case is normalized into one of twelve categories: Personnel Dispute, Pension Insurance Treatment Dispute, Labor Dispatch Contract Dispute, Medical Insurance Treatment Dispute, Unemployment Insurance Treatment Dispute, Work Injury Insurance Treatment Dispute, Maternity Insurance Treatment Dispute, Employment Relationship Confirmation Dispute, Welfare Treatment Dispute, Non-compete Clause Dispute, Economic Compensation Dispute, and Wage Recovery Dispute. For the balanced benchmark split, we sample 50 cases per category (600 total; see Table~\ref{tab:benchmark_distribution}).

Each instance is formatted as a triplet—\textbf{Instruction}, \textbf{Question}, and \textbf{Answer}—following the supervised fine-tuning (SFT) convention:
\begin{itemize}
    \item \textbf{Instruction:} Set the role as a legal expert in labor and social security law and classify the case based on the ``Court Findings.'' Output \emph{exactly one} label between \texttt{[Category]} and \texttt{<eoa>}, e.g., \texttt{[Category] Wage Recovery Dispute<eoa>}.
    \item \textbf{Question:} Provide the fact description excerpted from the ``Court Findings'' section. Each case is associated with one and only one category.
    \item \textbf{Answer:} Provide the gold label in the standardized format \texttt{[Category] <type><eoa>}.
\end{itemize}

A formatted instance is provided in Table~\ref{tab:t4_example}.

\begin{table}[!h]
\centering
\caption{Instruction format and an example for the Cause of Action Prediction task (T4).}
\label{tab:t4_example}
\renewcommand{\arraystretch}{1.25}
\setlength{\arrayrulewidth}{0.4pt}
\begin{tabular}{>{\columncolor{softgreen}}p{0.97\linewidth}}
\hline
\textbf{INSTRUCTION}: You are a legal expert in the field of labor and social security law. Determine the category of the cause of action found by the court in the text below. Each case belongs to exactly one category. The categories include: disputes over confirmation of labor relationship, labor dispatch contract disputes, wage claims, severance (economic compensation) disputes, non-compete disputes, pension insurance benefit disputes, work-injury insurance benefit disputes, medical insurance benefit disputes, maternity insurance benefit disputes, unemployment insurance benefit disputes, welfare benefit disputes, and personnel disputes. Write your answer strictly between \texttt{[Category]} and \texttt{<eoa>}, for example: \texttt{[Category] Disputes over confirmation of labor relationship<eoa>}. You must follow this format exactly.\\
\hline
\textbf{QUESTION}: Upon trial, it was ascertained that on March 4, 2020, the plaintiff and a third party, Kunshan Qianhong Enterprise Management Co., Ltd.\ (hereinafter ``Qianhong''), signed a \textit{Labor Dispatch Employment Contract}, agreeing that Qianhong would dispatch the plaintiff to Shishuo Electronics (Kunshan) Co., Ltd.\ (the user enterprise; hereinafter ``Shishuo Electronics'') to work as an operator. The dispatch period was from March 4, 2020 to September 4, 2020. The plaintiff worked at Shishuo Electronics until July 16, 2020. The plaintiff claimed that the defendant Shunxinren Company posted information in a WeChat group stating that one mode of working at Shishuo Electronics was: clocking in for 55 days and remaining employed for 90 days, for which a rebate of RMB 11{,}500 would be paid; the plaintiff provided WeChat chat records as evidence. On March 4, 2020, the defendant Shunxinren Company issued to the plaintiff a \textit{New Employee On-the-Job Reward Assistance Letter}, which stated that Chen Yichuan joined Shishuo Electronics through the defendant company and that the defendant promised to reward the employee RMB 7{,}000 after the employee had worked at the enterprise for 55 days and 550 working hours. On June 18, 2020, Shishuo Electronics paid the plaintiff a subsidy of RMB 4{,}500. The above facts are supported by the New Employee On-the-Job Reward Assistance Letter, the Labor Dispatch Employment Contract, transfer records, WeChat chat records, work permit, dispatch return approval form, absenteeism make-up registration, and the trial transcript of this court.\\
\hline
\textbf{ANSWER}: \texttt{[Category] Welfare benefit disputes<eoa>} \\
\hline
\end{tabular}
\end{table}

We report \textit{Accuracy} as the primary evaluation metric (see Table~\ref{tab:legal_tasks}). The balanced benchmark split contains 600 instances (50 per category; Table~\ref{tab:benchmark_distribution})

\subsection{Welfare Compensation Prediction}

To evaluate large language models' ability to handle welfare-related compensation and allowances in labor disputes, we design a prediction task (T5 in Table~\ref{tab:legal_tasks}) that focuses on prediction of welfare compensation, constructed from labor and social security judgments publicly released in Jiangsu Province over the past five years. Each judgment is structurally parsed to extract the plaintiff's claims, court findings, reasoning, and judgment; the “Court Findings” section serves as the model input.

The taxonomy of welfare compensation categories was systematically compiled by professional lawyers under the guidance of relevant laws and policies. We curate a canonical vocabulary of \emph{40+} categories with synonym normalization, including but not limited to: heatstroke prevention allowance, high-temperature allowance, hardship allowance, food subsidy, night-shift subsidy, only-child allowance, maternity allowance, funeral subsidy, childcare support, dependent rescue subsidy, relocation compensation, medical treatment subsidy, inpatient relief fund, transportation allowance, accommodation cost, rehabilitation cost, assistive device expenses, salary for suspension period, living assistance, disability allowance, retirement allowance, work-injury medical subsidy, disability resettlement subsidy, childcare funeral support, unused-leave wages, home-visit allowance, non-compete compensation, severance pay, settlement compensation, medical aid, penalty compensation, and overtime pay.

Each instance is uniformly cast into a supervised fine-tuning (SFT) triplet—\textbf{Instruction}, \textbf{Question}, and \textbf{Answer}:
\begin{itemize}
    \item \textbf{Instruction:} Set the model's role as a “legal expert in labor and social security law.” Require it to identify, from a predefined category list, the compensation/allowance items relevant to the case, and output strictly in the specified format.
    \item \textbf{Question:} Provide the factual description extracted from the judgment's “Court Findings” section to supply key information for identification.
    \item \textbf{Answer:} List the compensation/allowance categories actually involved in the case. When multiple items are present, concatenate them as a single list using the Chinese enumeration comma as the separator. Output strictly in the format \texttt{[Category] item1 item2 <eoa>}.
\end{itemize}

An example formatted instance is provided in Table~\ref{tab:t5_example}.

\begin{table}[!h]
\centering
\caption{Example of Welfare Compensation Prediction Task.}
\label{tab:t5_example}
\renewcommand{\arraystretch}{1.25}
\begin{tabular}{>{\columncolor{softgreen}}p{0.97\linewidth}}
\hline
\textbf{INSTRUCTION}: Based on the facts ascertained by the court in the text below, predict the welfare/compensation category(ies) relevant to this case. The categories include: heatstroke prevention and cooling allowance, high-temperature allowance, underground hardship position allowance, underground allowance, meal subsidy during the work shift, night-shift allowance, only-child allowance, maternity allowance, funeral subsidy, bereavement allowance, relief allowance for dependent relatives, relocation allowance for employees assigned to work in another locality, medical expenses, meal subsidy during hospitalization, transportation expenses, board and lodging expenses, rehabilitation treatment expenses, expenses for assistive devices for disability, wages during the medical treatment period with pay (work suspension with pay), living care expenses, one-time disability subsidy, disability allowance, retirement benefits, one-time work-injury medical subsidy, one-time disability employment subsidy, funeral subsidy, survivors' pension for dependent relatives, one-time death-in-service subsidy, wages for unused annual leave, travel expenses for home leave, non-compete compensation, liquidated damages for breach of a non-compete, economic compensation, payment in lieu of notice, compensation (damages), medical subsidy, liquidated damages, damages for breach, compensation for losses, and overtime pay. Write your answer strictly between \texttt{[Category]} and \texttt{<eoa>}, for example: \texttt{[Category] Heatstroke prevention and cooling allowance<eoa>}. You must follow this format exactly.\\
\hline
\textbf{QUESTION}: The court found as follows: Upon trial, this court finds the facts as follows. The plaintiff, Liyang Zheng, joined the defendant, Welfs Company, on March 24, 2016 to engage in spray-painting work. On May 29, 2017, she applied to resign for personal reasons, and rejoined the company the following month to resume the same work. The term of the last labor contract was from July 1, 2019 to June 30, 2021. On October 2, 2020, Liyang Zheng mailed to Welfs Company a notice of termination of the labor contract, stating that she had established a labor relationship with the employer on March 24, 2016; that her position was spray painting, which is a toxic and hazardous type of work; that the employer did not provide any labor-protection equipment as required, did not provide any medical examinations, and did not pay labor remuneration in a timely and full manner; and that she therefore terminated the labor relationship pursuant to Article 38 of the Labor Contract Law. Welfs Company signed for receipt the next day. On October 12 of the same year, Liyang Zheng applied for arbitration with the arbitration commission; the arbitration claims were the same as the claims in this lawsuit. The arbitration commission terminated the proceedings in the case by the arbitration decision Ningning Labor-Personnel Arbitration Decision No.\ (2020) 4983. Liyang Zheng brought the lawsuit before this court within the statutory time limit. Welfs Company submitted the labor contract and payroll records to prove that wages had been paid in full; submitted an environmental protection assessment report and warehouse-out records for labor-protection supplies to prove that it provided a safe working environment and lawfully provided labor-protection conditions; and submitted summer temperature statistics tables for 2020 and 2021.\\
\hline
\textbf{ANSWER}:\texttt{[Category] High-temperature allowance, economic compensation, overtime pay<eoa>}\\
\hline
\end{tabular}
\end{table}

For the balanced benchmark split, we select 50 representative samples per cause category (600 total; see Table~\ref{tab:benchmark_distribution}). We report \textit{F1} as the primary evaluation metric for this multi-label task (micro-F1 as the main score; macro-F1 as a secondary summary; see Table~\ref{tab:legal_tasks}).

\subsection{Named Entity Recognition (NER)}

To evaluate large language models' ability in entity recognition within labor law disputes, we design a structured extraction task (T6 in Table~\ref{tab:legal_tasks}) in the form of Named Entity Recognition, targeting three entity types: (1) \textit{Employee}, (2) \textit{Contracting Employer}, and (3) \textit{Actual Employer} . In labor disputes, three core parties are typically involved: the Employee, namely the natural person who provides labor and enters into the employment relationship under labor law; the Contracting Employer, referring to the contracting entity that signs the labor contract and assumes statutory obligations arising from the employment relationship; and the Actual Employer, which denotes the host entity where the worker actually performs services, even if no labor contract is directly concluded with that entity.

Each case was annotated by trained annotators and adjudicated for consensus. If an entity is not present, its value is set to an empty string. We standardize organization names to their registered forms (e.g., keep legal suffixes such as “Co., Ltd.”), normalize whitespace and punctuation, and preserve person names as written in the judgment.

Each sample follows the supervised fine-tuning triplet—\textbf{Instruction}, \textbf{Question}, and \textbf{Answer}:
\begin{itemize}
    \item \textbf{Instruction:} Act as a legal expert in labor and social security law. Extract the three entities from the “Court Findings” and output a JSON-like dictionary wrapped by \texttt{[Entity Recognition]} and \texttt{<eoa>}. \textbf{Use the fixed key order}: \texttt{Employee}, \texttt{Contracting Employer}, \texttt{Actual Employer}. If an entity is missing, return an empty string. Do not invent fields or values.
    \item \textbf{Question:} Provide the “Court Findings” excerpt from the judgment as the model input.
    \item \textbf{Answer:} Return the dictionary in the standardized format [Entity Recognition] \{Employee: 'A', Contracting Employer: 'B', Actual Employer: 'C'\}<eoa>.
\end{itemize}

A formatted instance is provided in Table~\ref{tab:t6_example}.

\begin{table}[!h]
\centering
\caption{Instruction format and an example for the Named Entity Recognition task (T6).}
\label{tab:t6_example}
\renewcommand{\arraystretch}{1.25}
\setlength{\arrayrulewidth}{0.4pt}
\begin{tabular}{>{\columncolor{softgreen}}p{0.97\linewidth}}
\hline
\textbf{INSTRUCTION}: You are a legal expert in the field of labor and social security law. Based on the facts ascertained by the court in the text below, extract named entities. The entity types include: \textit{Worker}, \textit{Employer}, and \textit{User Enterprise}. Return a dictionary listing the extracted entities, and write your answer strictly between \texttt{[NER]} and \texttt{<eoa>}, for example: \texttt{[NER]\{'Worker': 'Zhou Yingyu', 'Employer': 'Huangdai Middle School', 'User Enterprise': 'Suzhou Dinghao Electric Appliance Co., Ltd.'\}<eoa>}. If an entity type is truly absent, use an empty string as the corresponding value, for example: \texttt{[NER]\{'Worker': 'Hu Zhongnan', 'Employer': 'Baifu Co., Ltd.', 'User Enterprise': ''\}<eoa>}. You must follow this format exactly.\\
\hline
\textbf{QUESTION}: The court found: On September 26, 2017, Chengnuo Company recruited Li (Party B) and then dispatched him to Henghua Company to work as a binding worker. Neither Chengnuo Company nor Henghua Company paid work-injury insurance or other social insurance contributions for Li (Party B). On November 30, 2018, Li (Party B) worked from 20:30 to 08:30 the next day. At 05:30 on December 1, while at work and on his way to the restroom, he was found by a coworker unconscious in the restroom; despite rescue efforts, he died later that day. Cause of death: (1) myocardial infarction; (2) cerebrovascular accident. After the incident, Li (Party A) and others signed a \textit{Work-Injury Death Compensation Agreement} with Henghua Company, agreeing that Henghua Company would pay work-injury death benefits of RMB 420{,}000 for Li (Party B). Thereafter, Li (Party A) and others shall not make any form of compensation claim against Henghua Company regarding death-in-service compensation for Li (Party B) or any rights Li (Party B) enjoyed during his dispatched work at Henghua Company, and shall not engage in disturbances, petitions, etc. Li (Party A) and others shall cooperate with Henghua Company in applying for work-injury determination or labor arbitration, and cooperate with Henghua Company in any necessary litigation or recourse against Chengnuo Company. On December 28, 2018, the Dantu District Human Resources and Social Security Bureau of Zhenjiang City issued a work-injury determination decision, determining that the accidental injury suffered by Li (Party B) fell within the scope of work-injury determination and was recognized as a work injury. Subsequently, Li (Party A) and others filed a case against Chengnuo Company regarding work-injury insurance benefits disputes with the Dantu District Labor and Personnel Dispute Arbitration Commission of Zhenjiang City. Both Li (Party A) and Chengnuo Company were dissatisfied with the arbitral award and brought the case to court. Both parties argued that Henghua Company should bear joint and several liability with Chengnuo Company for the death-in-service compensation for Li (Party B); additionally, Chengnuo Company argued that the RMB 420{,}000 already paid by Henghua Company should be deducted.\\
\hline
\textbf{ANSWER}:\texttt{[NER]\{'Worker': 'Li (Party B)', 'Employer': 'Chengnuo Company', 'User Enterprise': 'Henghua Company'\}<eoa>} \\
\hline
\end{tabular}
\end{table}

The \textbf{balanced benchmark} split contains 600 instances (50 per dispute category; see Table~\ref{tab:benchmark_distribution}). We report \textit{Soft-F1} as the primary metric (micro-averaged over the three fields, with punctuation/whitespace normalization), and provide exact-match F1 as a secondary reference (Table~\ref{tab:legal_tasks}).

\subsection{Claim Mining}

This task focuses on mining claims made by plaintiffs in labor dispute cases. 
The dataset is constructed from publicly available labor-related court decisions in Jiangsu Province over the past five years. 
From each judgment document, we use the “Facts and Reasoning” section as model input, guiding the model to identify and extract the plaintiff's claims. 
The ground-truth labels were derived from the “Claims” section of the judgment, reformulated and standardized by legal experts to ensure clarity, conciseness, and structural consistency.

Each data instance is represented in the form of a triplet—\textbf{instruction}, \textbf{question}, and \textbf{answer}—defined as follows:

\begin{itemize}
    \item \textbf{Instruction:} Instructs the model to assume the role of a legal expert in labor and social security law. The model is prompted to \textbf{mine and extract the plaintiff's claims} based on the “Facts and Reasoning” section.
    \item \textbf{Question:} Provides the fact description (i.e., the “Facts and Reasoning” segment) from the judgment as the model input.
    \item \textbf{Answer:} Contains the reformulated original claims made by the plaintiff, as recorded in the “Claims” section of the judgment.
\end{itemize}

An example of this task is shown below table~\ref{tab:t7_example}:

\begin{table}[!h]
\centering
\caption{Example of Claim Mining Task.}
\label{tab:t7_example}
\renewcommand{\arraystretch}{1.25}
\begin{tabular}{>{\columncolor{softgreen}}p{0.97\linewidth}}
\hline
\textbf{INSTRUCTION}: You are a legal expert in the field of labor and social security law. Based on the appellant's stated facts and grounds below, summarize the litigation claim(s) being requested.\\
\hline
\textbf{QUESTION}: The facts and grounds are as follows. The father of the two plaintiffs, Pan Jingliang, was a retired employee of the defendant entity during his lifetime. Throughout his lifetime, his retirement wages were paid by the defendant, and he signed an agreement with the defendant providing that funeral expenses would be paid after his death; however, the agreement did not provide for a one-time survivor benefit payable to his lineal relatives after his death. According to the Jiangsu Provincial Department of Human Resources and Social Security's \textit{Notice on Adjusting Benefits for Enterprise Employees and Retirees Who Die of Illness or Non-Work-Related Causes}, after an enterprise employee dies, a one-time survivor benefit of RMB 16{,}000 should be paid to the decedent's lineal relatives.\\
\hline
\textbf{ANSWER}: \textit{Order the defendant to pay a survivor benefit of RMB 16{,}000 in connection with the death of Pan Jingliang.}\\
\hline
\end{tabular}
\end{table}

In total, we collected 600 samples from real-world court documents. These were categorized into 12 major case types. From each category, 50 representative samples were selected, resulting in a benchmark dataset of 600 items for this task.

\subsection{Disputed Issue Mining}

This task aims to extract the core dispute issues between the parties in labor-related legal cases. This task is constructed from labor and social security court decisions publicly released in Jiangsu Province over the past five years. The claims made by both the plaintiff and the defendant in the judgment documents are used as the input, while the court-identified disputed issues, summarized on the basis of the parties' arguments, serve as the ground-truth answers.

Each data instance is structured as a triplet—\textbf{instruction}, \textbf{question}, and \textbf{answer}—as described below:

\begin{itemize}
\item \textbf{Instruction:} Instructs the model to act as a legal expert in labor and social security law. The model is asked to identify and summarize the isputed issue mining based on the claims made by both parties.
\item \textbf{Question:} Provides the core arguments made by both the plaintiff and defendant, extracted from the judgment. This section includes the background of the dispute, claims and counterclaims, and key facts, serving as the basis for identifying the disputed issue.
\item \textbf{Answer:} Contains the court's summarized focus of the dispute, as identified in the judgment.
\end{itemize}

In total, we compiled 600 samples, covering 12 major dispute categories: Personnel Disputes (50), Pension Insurance (50), Labor Dispatch Contracts (50), Medical Insurance (50), Unemployment Insurance (50), Work Injury Insurance (50), Maternity Insurance (50), Employment Relationship Confirmation (50), Welfare Treatment (50), Non-compete Clause (50), Economic Compensation (50), and Wage Recovery (50). These examples jointly form the dataset used for this task. An example of this task is shown below table~\ref{tab:t8_example}:

\begin{table}[!h]
\centering
\caption{Example of Dispute Focus Extraction Task.}
\label{tab:t8_example}
\renewcommand{\arraystretch}{1.25}
\begin{tabular}{>{\columncolor{softgreen}}p{0.97\linewidth}}
\hline
\textbf{INSTRUCTION}: You are a legal expert in the field of labor and social security law. Based on the parties' claims and arguments in the text below, summarize the key disputed issues in this case.\\
\hline
\textbf{QUESTION}: The plaintiff (defendant), Hongsheng Company, submitted the following claims to this court: (1) order that Hongsheng Company is not required to pay funeral subsidies of RMB 25{,}184; (2) order that Hongsheng Company is not required to pay dependent relatives' survivors' pension of RMB 54{,}000 to Fan Wenguang; and (3) litigation costs to be borne by the other party. Facts and grounds: The arbitration award ordering Hongsheng Company to pay the above amounts is improper because the death was caused by a third party's tort, and the family has already received corresponding compensation; under the \textit{Jiangsu High People's Court Labor Dispute Trial Guide} and the \textit{Guiding Opinions of the Jiangsu High People's Court on Properly Trying Labor Dispute Cases Under the Current Macroeconomic Situation}, such compensation should be deducted. In addition, if pension insurance benefits have already been received, survivors' benefits should not be paid. The defendant (plaintiff), Fan Wenguang, Wang Cuiying, Fan Zhengjie, Fan Yujuan, and Fan Yulian, submitted the following claims to this court: (1) require Hongsheng Company to pay funeral subsidies, dependent relatives' survivors' pension, and a one-time death-in-service subsidy, totaling RMB 1{,}346{,}962; and (2) litigation costs to be borne by the other party. Facts and grounds: The arbitration award incorrectly applied the standard for the ``previous year'' as the year preceding the accident (i.e., 2014); it should instead be the year preceding the conclusion of court debate. The arbitration award's support for five years of survivors' pension for Fan Wenguang is improper: Fan Wenguang has already survived for four years and did not receive survivors' pension during that period; under the judicial interpretation based on ``the location of the court accepting the case,'' awarding five years is clearly improper. Wang Cuiying and Fan Zhengjie have limited work capacity and should be granted a certain amount of survivors' benefits.\\
\hline
\textbf{ANSWER}: \textit{Disputed issues: (1) the applicable standard for the one-time death-in-service subsidy; (2) whether funeral subsidies should be paid; and (3) whether dependent relatives' survivors' benefits should be paid, and the applicable standard.}\\
\hline
\end{tabular}
\end{table}

\subsection{Statute Prediction (Based on Facts)}
This task focuses on predicting the statutes referenced by the court based on the factual description of a labor dispute case. The dataset is constructed from publicly available labor-related court decisions in Jiangsu Province over the past five years. Specifically, the “Court Findings” section is extracted as model input, and the model is required to infer the legal provisions cited by the court based on the factual content. The ground-truth labels are derived from the legal provisions explicitly cited in the “Judgment Result” or “Court's View” sections of the judgment.

Each data instance is represented in the form of a triplet—\textbf{instruction}, \textbf{question}, and \textbf{answer}—defined as follows:

\begin{itemize}
\item \textbf{Instruction:} Instructs the model to assume the role of a legal expert in labor and social security law. The model is prompted to identify relevant statutes and specific clauses based on the factual content in the “Court Findings” section.
\item \textbf{Question:} Provides the factual description of the case, as stated in the “Court Findings” section, including details on background, actions, contract execution, and disputes.
\item \textbf{Answer:} Contains the statute(s) cited by the court, including their full text and clause numbers, as referenced during the adjudication process.
\end{itemize}

In total, we curated 600 statute prediction samples from real-world court documents, spanning 12 case types. From each category, 50 representative samples were selected, resulting in a benchmark dataset of 600 items for this task. An example of this task is shown below~\ref{tab:t9_example}:

\begin{table}[!t]
\centering
\caption{Example of Statute Prediction (Based on Facts).}
\label{tab:t9_example}
\scriptsize
\renewcommand{\arraystretch}{1.08}
\setlength{\tabcolsep}{3pt}

\begin{tabular}{>{\columncolor{softgreen}}p{0.97\linewidth}}
\hline
\textbf{INSTRUCTION}: You are a legal expert in labor and social security law. Based on the court-found facts, provide relevant statutory provisions and key contents (one or more may apply).\\
\hline
\textbf{QUESTION}: Court-found facts (abridged): Fu Yunshan suffered a work-related traffic injury while employed by Huayuan Company (Aug.\ 2006--Jun.\ 2010). Work injury was recognized (Aug.\ 30, 2011) and assessed as Grade 5 disability. Multiple arbitration attempts on work-injury compensation were terminated; Fu Yunshan died on Dec.\ 9, 2013, and relatives participated as plaintiffs.\\
\hline
\textbf{ANSWER} (key provisions): 
\begin{itemize}\setlength{\itemsep}{1pt}\setlength{\parskip}{0pt}\setlength{\topsep}{1pt}
  \item \textit{Regulations on Work-Related Injury Insurance} (2010), Art.\ 36: Grade-5 benefits include one-time disability subsidy (18 months' wage) and monthly disability allowance (70\% wage) with social insurance contributions; additional one-time medical/employment subsidies upon termination (local standards apply).
  \item \textit{Jiangsu Implementing Measures} (2015), Arts.\ 22 \& 24: employer duty for timely treatment; rehabilitation plan and approved rehabilitation services via contracted institutions.
  \item \textit{Civil Procedure Law} (2012), Art.\ 64: burden of proof; court may investigate/collect evidence when necessary.
\end{itemize}
\\
\hline
\end{tabular}
\end{table}

\subsection{Statute Prediction (Based on Scenarios)}
This task is built upon statutory provisions from the \emph{Labor Law} and related regulations that contain clearly defined conditions of applicability. 
We construct a fact-pattern oriented statute prediction dataset designed to cover common labor dispute scenarios. 
Specifically, each sample takes a statutory provision as the gold label, and a large language model (e.g., GPT) generates a realistic labor dispute scenario consistent with that provision. 
Each instance is limited to a single statutory clause to ensure that the generated scenario points unambiguously to a specific article. 
Given the hypothetical scenario and corresponding question, the model must accurately output the applicable statute along with its full textual content.

Each data instance is represented in the form of a triplet—\textbf{instruction}, \textbf{question}, and \textbf{answer}—defined as follows:

\begin{itemize}
\item \textbf{Instruction:} Instructs the model to assume the role of a legal expert in labor and social security law. The model is prompted to determine the applicable statutory provision based on the scenario description and return its full text.
\item \textbf{Question:} A hypothetical dispute scenario generated with reference to a specific statutory provision, representing the factual setting for legal application.
\item \textbf{Answer:} The statutory provision applicable to the scenario, including its article number and full text.
\end{itemize}

An example of this task is shown table~\ref{tab:t10_example}:

\begin{table}[!t]
\centering
\caption{Example of Statute Prediction (Based on Scenarios).}
\label{tab:t10_example}
\renewcommand{\arraystretch}{1.25}
\begin{tabular}{>{\columncolor{softgreen}}p{0.97\linewidth}}
\hline
\textbf{INSTRUCTION}: You are a legal expert in the field of labor and social security law. Based on the hypothetical scenario and question below, provide the legal basis. You only need to provide the relevant statutory provision and its specific text. Each scenario involves exactly one provision.\\
\hline
\textbf{QUESTION}: Hypothetical scenario: A construction company subcontracts a project to a labor contractor who lacks lawful business qualifications, and as a result, the contractor defaults on wages owed to the migrant workers under him. In this situation, how should the migrant workers protect their rights and interests? Under which legal provision, and who should ultimately be responsible for paying the migrant workers' wages?\\
\hline
\textbf{ANSWER}: Statutory provision: \textit{Regulations on Ensuring the Payment of Wages to Migrant Workers}, Article 19: Where an employer subcontracts work tasks to an individual or to an entity that lacks lawful business qualifications, resulting in arrears of wages owed to the migrant workers it recruits, the matter shall be handled in accordance with relevant laws and regulations. Where an employer allows an individual, an entity that lacks lawful business qualifications, or an entity that has not obtained the corresponding qualifications to conduct business externally in the name of the employer, resulting in arrears of wages owed to the migrant workers it recruits, the employer shall pay off the arrears and may seek recourse in accordance with law.\\
\hline
\end{tabular}
\end{table}

\subsection{Case Analysis (Without Statutes)}

This task focuses on legal reasoning in the absence of explicitly cited statutes. The dataset was constructed from labor-related court decisions publicly released in Jiangsu Province over the past five years. Specifically, we extracted the “Court's Reasoning” section from each judgment as input, requiring the model to conduct a three-step legal analysis based on the facts presented. The ground-truth labels were derived from the “Court's Opinion” section, where the court applies legal logic without referencing any statutory provision, serving as the gold-standard answer for evaluating the model's reasoning adequacy.

Each data instance is structured as a triplet—\textbf{instruction}, \textbf{question}, and \textbf{answer}—as defined below:

\begin{itemize}
\item \textbf{Instruction:} Instructs the model to assume the role of a legal expert in labor and social security law and to perform a three-step legal reasoning process based on the case facts provided, drawing logical inferences without citing explicit legal statutes.
\item \textbf{Question:} Presents the factual context, extracted from the “Court's Reasoning” section of the judgment, as the basis for the model to infer legal conclusions.
\item \textbf{Answer:} Contains the final judgment reasoning extracted from the “Court's Opinion” section, where the court applies practical legal logic to reach a conclusion without referencing specific statutory articles.
\end{itemize}

An example of this task is shown below  table\ref{case_analysis}:

\begin{table}[!t]
\centering
\caption{Example of Case Analysis (Without Statutes).}
\label{case_analysis}
\renewcommand{\arraystretch}{1.25}
\begin{tabular}{>{\columncolor{softgreen}}p{0.97\linewidth}}
\hline
\textbf{INSTRUCTION}: You are a legal expert in the field of labor and social security law. Based on the court-found facts below, conduct a three-step syllogistic analysis of the disputed issue(s).\\
\hline
\textbf{QUESTION}: The court found: The defendant, Yidajia Company, is an operator of a \textit{Ele.me} delivery station. After being recruited by Niu (a station manager of the company), the plaintiff, Jiang (a delivery rider), registered on the \textit{Ele.me} app platform and accepted delivery orders exclusively through that platform. Jiang's daily work was arranged and managed by the station manager; his working hours were from 7:00 a.m.\ to 7:00 p.m.; and Yidajia Company paid him wages on a monthly basis. Later, the station manager requested that Jiang leave the job on the grounds that he refused to accept orders and made late deliveries. After unsuccessful communication with Yidajia Company, Jiang applied for labor arbitration, requesting confirmation that a labor relationship existed between him and Yidajia Company and seeking compensation for unlawful termination of the labor contract. Jiang then filed suit after he disagreed with the arbitral award. During trial, Yidajia Company submitted a \textit{Delivery Service Contracting Agreement}, which provided, inter alia: ``Jiang entrusts Yidajia Company to handle industrial and commercial registration to establish an individual business entity, and the food-delivery service shall be contracted by Jiang \ldots''\\
\hline
\textbf{ANSWER}: \textit{Major premise (rule).} Article 1 of the \textit{Notice on Matters Concerning the Establishment of Labor Relationships} (Ministry of Labor and Social Security of the PRC, 2005, No.\ 12) provides that where an employer recruits a worker without entering into a written labor contract, a labor relationship is established if all of the following are satisfied: (1) the employer and the worker have the subject qualifications as required by laws and regulations; (2) the employer's labor rules and regulations formulated in accordance with law apply to the worker, the worker is subject to the employer's labor management, and the worker performs remunerated work arranged by the employer; and (3) the labor provided by the worker constitutes a component of the employer's business.\\[4pt]
\textit{Minor premise (facts).} In this case, although the agreement states that Jiang must establish an individual business entity to ``contract'' food-delivery services, the existence of a labor relationship should be determined based on the actual circumstances. Jiang was recruited to work at the defendant's station; his work was arranged and managed by the station manager, who acted on behalf of the station operator; Jiang worked fixed hours (7:00 a.m.\ to 7:00 p.m.); and his wages were assessed and paid monthly by Yidajia Company.\\[4pt]
\textit{Conclusion.} The above facts satisfy the legal characteristics for establishing a labor relationship under the Notice. Therefore, a labor relationship exists between the plaintiff Jiang and the defendant Yidajia Company.\\
\hline
\end{tabular}
\end{table}

In total, we collected 384 representative samples across 12 legal categories: Personnel Disputes (32), Pension Insurance (32), Labor Dispatch Contracts (32), Medical Insurance (32), Unemployment Insurance (32), Work Injury Insurance (32), Maternity Insurance (32), Employment Relationship Confirmation (32), Welfare Benefits (32), Non-compete Clauses (32), Economic Compensation (32), and Wage Recovery (32). These items collectively constitute the benchmark dataset for this task.

\subsection{Case Analysis (With Statutes)}

This task focuses on analyzing labor dispute cases based on statutes. The dataset was constructed from publicly available labor-related court decisions in Jiangsu Province over the past five years. From each judgment document, we extracted the “Court Inquiry” section as model input, and required the model to conduct a statutory analysis focused on applying relevant laws to the case. Standard answers were derived from the “Court's Judgment” section of the judgment, specifically the parts referring to relevant statutes, ensuring clarity and logical consistency in their application.

Each data instance is represented in the form of a triplet—\textbf{instruction}, \textbf{question}, and \textbf{answer}—defined as follows:

\begin{itemize}
    \item \textbf{Instruction:} Instructs the model to assume the role of a legal expert in labor and social security law. The model is prompted to analyze the case based on the “Court Inquiry” section, using relevant statutes for the legal application.
    \item \textbf{Question:} Provides the fact description (i.e., the “Court Inquiry” segment) from the judgment as the basis for legal inference, including the case's background, actions of the parties, and the dispute.
    \item \textbf{Answer:} Contains the standard analysis based on relevant statutes cited in the judgment, including logical analysis and conclusions derived from applying these laws.
\end{itemize}

In total, we collected 384 representative samples across 12 legal categories: Personnel Disputes (32), Pension Insurance (32), Labor Dispatch Contracts (32), Medical Insurance (32), Unemployment Insurance (32), Work Injury Insurance (32), Maternity Insurance (32), Employment Relationship Confirmation (32), Welfare Benefits (32), Non-compete Clauses (32), Economic Compensation (32), and Wage Recovery (32). These items collectively constitute the benchmark dataset for this task.
An example of this task is shown below table \ref{case_analysis2}:

\begin{table}[!t]
\centering
\caption{Example of Case Analysis (With Statutes).}
\label{case_analysis2}
\renewcommand{\arraystretch}{1.20}
\begin{tabular}{>{\columncolor{softgreen}}p{0.97\linewidth}}
\hline
\textbf{INSTRUCTION}: You are a legal expert in labor and social security law. Based on the court-found facts and the referenced provision, conduct a three-step syllogistic analysis of the disputed issue(s).\\
\hline
\textbf{QUESTION}: A construction company subcontracted a project to Mr.\ Liu, who hired Mr.\ Zhang in Aug.\ 2021. On Oct.\ 10, 2021, Mr.\ Zhang fell at the site and suffered a lumbar vertebral fracture. A prior judgment confirmed no labor relationship between Mr.\ Zhang and the company. In Mar.\ 2023, the authority recognized the injury as work-related and determined the company should bear work-injury insurance liability. Mr.\ Zhang was assessed as Grade 8 disability, with a 6-month paid suspension. He sought Grade 8 benefits via arbitration; the company sued after losing.\\[2pt]
\textit{Disputed issue:} Whether Mr.\ Zhang is entitled to work-related injury insurance benefits.\\[2pt]
\textit{Referenced provision:} Article 3(4) of the \textit{SPC Provisions on Administrative Cases of Work-Related Injury Insurance} provides that where a user enterprise illegally subcontracts contracted business to an unqualified organization or natural person, and an employee hired thereby suffers injury or death arising from work while performing the contracted business, the user enterprise shall bear work-related injury insurance liability.\\
\hline
\textbf{ANSWER}: \textit{Major premise (rule).} Under Article 3(4), work-injury insurance liability is not conditioned on a labor relationship. If a qualified user enterprise illegally subcontracts to an unqualified entity/person and the hired worker is injured at work while performing the subcontracted business, the user enterprise bears liability.\\[3pt]
\textit{Minor premise (facts).} The company subcontracted the project to Mr.\ Liu, who hired Mr.\ Zhang to work on-site. Mr.\ Zhang was injured while performing the work, and the authority recognized it as work-related. Although no labor relationship exists between Mr.\ Zhang and the company, the facts fall within Article 3(4).\\[3pt]
\textit{Conclusion.} Therefore, the company should bear work-injury insurance liability for Mr.\ Zhang and pay the corresponding benefits (including Grade 8 benefits) if contributions were not paid.\\
\hline
\end{tabular}
\end{table}

\section{Domain-Specialized Labor Law Model}
To obtain a strong labor law–specific model on top of a general-purpose backbone, we construct \textbf{LaborLawLLM} by supervised fine-tuning Qwen2.5-7B \cite{qwen2023} on a curated instruction-tuning corpus derived from Chinese labor and social security law sources. The data construction follows the unified *Instruction–Question–Answer* format introduced in Section III, which has already been used for all twelve benchmark tasks (T1–T12) spanning the twelve labor-law case categories C1–C12. 

\textbf{Training data.} Beyond the benchmark splits used purely for evaluation, we additionally build \textbf{51,236} supervised fine-tuning (SFT) instances. Each instance is formatted as an instruction that assigns the role of a labor and social security law expert, a task-specific question (e.g., statute recall, examination-style Q\&A, case facts, or dispute descriptions), and a reference answer consistent with the task definitions in Section III (such as standardized option labels, statute text, or normalized court-style analyses). This SFT corpus is designed to cover the core skills required by LaborLawBench—statute memorization, exam-style doctrinal knowledge, case-type prediction, welfare and compensation reasoning, named entity recognition, and short- and long-form case analysis—so that the model is explicitly aligned with the task interfaces and output constraints used in our benchmark. 

\textbf{Base model and fine-tuning strategy.} We start from the open-source \textbf{Qwen2.5-7B} model, a 7B-parameter general Chinese large language model that serves as one of our main baselines in Section \ref{sec:experiment}.  To adapt it efficiently to the labor law domain, we adopt parameter-efficient fine-tuning with \textbf{LoRA}, rather than updating all base parameters. The model is trained for \textbf{8 epochs} over the 51,236-example SFT set. We use a \textbf{per-device batch size of 1} and \textbf{64 gradient accumulation steps}, on \textbf{4× NVIDIA RTX 4090} GPUs. This corresponds to an effective batch of 256 sequences per optimization step. The full fine-tuning run takes approximately \textbf{22 days}, reflecting the computational cost of repeatedly exposing the model to the diverse labor-law tasks and case types defined by our benchmark.

\textbf{Optimization details.} We follow the standard autoregressive supervised fine-tuning objective for causal LLMs: given the concatenated Instruction–Question–Answer sequence, the model is trained to maximize the likelihood of the gold answer tokens conditioned on the instruction and question. We apply the same text normalization and formatting rules used in the benchmark (e.g., constrained answer formats such as `[Correct Answer]X<eoa>` for examination questions and bracketed tags for structured outputs) so that the model learns to produce outputs that are both legally accurate and format-compatible with our evaluation scripts.

\section{Evaluaton Metric}

After obtaining the predictive results from the model, we compare the predictions against the ground truth provided in the dataset. The dataset includes instructions, questions, and corresponding answers for each data instance, forming a complete key–value pair. Through this, we evaluate the model's ability to predict with good data foundations. In the evaluation process, each task is mainly divided into two steps: (1) extract model-generated predictions for the questions and compare them to the model's answers and related questions from the dataset; (2) compute evaluation metrics based on the model's predictions, which help in assessing the model's prediction consistency and relevance. For each evaluation task, we use five different evaluation metrics to cover various types of tasks. The following details describe the overall process of how each metric is calculated.

As summarized in Table~\ref{tab:legal_tasks}, the benchmark comprises twelve tasks spanning generation, classification (single- and multi-label), extraction, and regression. Each task is paired with an appropriate metric and a fixed data size: ROUGE-L for legal provision generation and scenario-based provision prediction (T1, T10; 3{,}165 and 670 items, respectively), Accuracy for single- and multi-label knowledge questions and cause prediction (T2–T4; 620/310/600 items), micro-F1 for welfare compensation (T5; 600 items), Soft-F1 for named entity recognition (T6; 600 items), and a GPT-O1 judge for claim/dispute extraction, fact-based provision identification, and case analyses (T7–T9, T11–T12; 600/600/600 and 384/384 items). Unless otherwise noted, scores are averaged over all instances reported in Table~\ref{tab:legal_tasks}, and abstentions (unparsable or format-violating outputs) receive zero credit under the corresponding metric.

\subsection{ROUGE-L}
ROUGE-L is a standard evaluation metric used to measure the content alignment between the predicted answer and the reference answer. It calculates the longest common subsequence (LCS) between the two sequences of content, ensuring the model's predictions are evaluated by the alignment of content and structure. ROUGE-L is particularly suitable for tasks involving sequence generation, as it takes into account both content and order alignment. We use ROUGE-L as our main evaluation metric for two tasks:

For task T1, the metric is suitable for tasks where the model needs to refine and generate specific legal clauses based on certain rules, which corresponds to the alignment of legal texts. For task T10, which deals with model evaluation on prediction tasks, the metric is used to measure the prediction accuracy in terms of its alignment with the corresponding reference answers.

In the process of extracting answers from the keys that make up the question-answer pairs, we retrieve the answers by accessing the “prediction” key and processing the output to verify its consistency with the format of the standard answer. We clean the answer by removing any irrelevant characters such as quotation marks and extra symbols, ensuring consistency with the reference format.

ROUGE-L calculates the longest common subsequence (LCS) between the predicted answer and the standard answer, which measures the alignment of content and order between the two sequences. The formula for ROUGE-L is as follows:

\begin{equation}
ROUGE\text{-}L = \frac{(1 + \beta^2) \cdot R_{LCS} \cdot P_{LCS}}{\beta^2 \cdot R_{LCS} + P_{LCS}}
\end{equation}

where:

\begin{equation}
R_{LCS} = \frac{LCS(X, Y)}{length(Y)}, \quad P_{LCS} = \frac{LCS(X, Y)}{length(X)}
\end{equation}

\begin{itemize}
    \item $X$: predicted answer
    \item $Y$: standard answer
    \item $LCS(X, Y)$: the length of the longest common subsequence between $X$ and $Y$
    \item $length(X)$: the length of $X$
    \item $length(Y)$: the length of $Y$
    \item $\beta$: a weight parameter used to balance recall and precision, typically set to 1
\end{itemize}

In an evaluation example, we present a case from the evaluation task. The instruction is: "You are a legal expert in labor and social security law." The answer to the question is as follows:

\textbf{Instruction:} You are a legal expert in labor and social security law. Based on the following content from a labor dispute case, please summarize the first article of the labor law regarding the case.

\textbf{Question:} The court examined the case: "The Chinese People's Court and the International Labor Arbitration decided that the plaintiff should receive compensation for the work injury..." 

\textbf{Answer:} Based on the law, the compensation must be provided to the plaintiff for their injuries.

\noindent This example helps us evaluate the alignment and clarity of the question and answer.

For evaluation, we use the ROUGE-L metric to calculate the longest common subsequence (LCS) between the predicted answer and the standard answer. ROUGE-L measures the alignment of content and order between the two sequences.

The formula for ROUGE-L is as follows:

\begin{equation}
P = \frac{LCS \text{ length}}{\text{Predicted Answer Length}} \quad (\text{Precision})
\end{equation}

\begin{equation}
R = \frac{LCS \text{ length}}{\text{Standard Answer Length}} \quad (\text{Recall})
\end{equation}

\begin{equation}
F_1 = \frac{2 \cdot P \cdot R}{P + R} \quad (\text{F1 Score})
\end{equation}

In this example, the precision rate is 59.09\%, recall is 100\%, and the F1 score is 74.29\%.

balanced and reliable evaluation metric in the legal domain.

\subsection{Accuracy}

We evaluate single-label and multi-label classification tasks with \emph{Accuracy}, covering \textbf{T2} (single-choice knowledge Q\&A), \textbf{T3} (multiple-choice knowledge Q\&A), and \textbf{T4} (cause-of-action prediction). In single-label settings, the model must select the unique correct option among candidates; accuracy reflects its ability to identify the correct label. In multi-label settings, the gold answer may contain several correct options; we adopt an \emph{exact set match} criterion, i.e., the prediction must include all and only the correct options to be counted as correct.

For each instance, we extract the gold label(s) from the \texttt{answer} field and the model output from the \texttt{prediction} field. To ensure comparability, we normalize both sides by (i) removing superfluous prefixes/suffixes such as ``\texttt{[Correct Answer]}'' and ``\texttt{<eoa>}'', (ii) stripping quotation marks and line breaks, and (iii) collapsing whitespace and standardizing case (uppercasing option letters). The expected output format is \texttt{[Correct Answer] A <eoa>} for single-label and \texttt{[Correct Answer] A,C <eoa>} (comma-/space-separated) for multi-label. If the exact format is absent, we attempt to recover labels by matching against the admissible option set (e.g., \{A,B,C,D\}) within the free-form text. If no valid option can be recovered, or if recovered options are outside the admissible set, the prediction is deemed invalid.

Formally, let $\mathcal{D}=\{1,\ldots,N\}$ index evaluation instances, $\mathcal{C}$ be the option set, $Y_i\subseteq\mathcal{C}$ the gold label set for instance $i$ (with $|Y_i|=1$ in single-label tasks), and $\widehat{Y}_i\subseteq\mathcal{C}$ the extracted prediction. For single-label tasks, we require $|\widehat{Y}_i|=1$; otherwise the instance is counted as incorrect. For multi-label tasks, order is ignored and duplicates are removed during extraction. We report micro-accuracy as exact set match:
\begin{equation}
\text{Accuracy} \;=\; \frac{1}{N}\sum_{i=1}^{N} \mathbf{1}\!\left[\;\widehat{Y}_i = Y_i\;\right],
\end{equation}
where $\mathbf{1}[\cdot]$ is the indicator function. Invalid or empty predictions are scored as incorrect (contributing $0$ to the numerator).

\textit{Illustrative example (single-label, T2).} 
Instruction: ``You are a legal expert in labor and social security. Please select the correct answer from A, B, C, D and output in the format \texttt{[Correct Answer] Answer <eoa>} (e.g., \texttt{[Correct Answer] A <eoa>}).''  
Question: ``Company A signed a labor contract with Zhang on January~5, 2010, agreeing on a one-month probation period. On February~1, 2010, the contract was notarized. On March~1, 2010, Zhang reported to Company A. When is the labor relationship established?  
A. January 5, 2010 \quad
B. February 1, 2010 \quad
C. March 1, 2010 \quad
D. April 1, 2010''

Gold answer: \texttt{[Correct Answer] C <eoa>}.  
Model output: ``According to the Labor Contract Law, a labor relationship is established from the date the employee starts work. Therefore, the answer is: C. March 1, 2010.''  
Although the model did not use the canonical bracketed format, the post-processor recovers the valid option \texttt{C} from the admissible set $\{A,B,C,D\}$. Since $\widehat{Y}_i=\{C\}=Y_i$, the instance is counted as correct (contributes $1/N$ to accuracy). 

\textit{Implementation notes.} For single-label tasks, if multiple distinct options are detected in the prediction, the instance is marked incorrect. For multi-label tasks, we treat labels as an unordered set (e.g., ``\texttt{B,A}'' equals ``\texttt{A,B}''); predictions that are strict supersets or subsets of $Y_i$ are marked incorrect under the exact-match criterion. This strict treatment ensures that accuracy reflects precise coverage of all and only the legally correct options.

\subsection{F1 Score Evaluation}

F1 score is a widely adopted metric for classification tasks, defined as the harmonic mean of precision and recall. For the T5 welfare compensation classification task, we adopt the F1 score as the primary evaluation metric. The goal of this task is to accurately identify key categories of welfare compensation mentioned in legal texts while minimizing omissions. The F1 score effectively captures both the precision and coverage of model predictions, making it a suitable metric for evaluating this task.

To evaluate this task, we define a set of standard compensation categories (\texttt{option\_list}) as the possible correct answers. In each sample, we require the model to produce output in the format ``\texttt{Category} \texttt{Answer} <eoa>" under the instruction field. During answer extraction, we first check whether the answer text contains the pattern ``\texttt{Category} \texttt{Answer} <eoa>". If found, we extract the content as the true answer category; if not, the entire answer text is used as the answer content. We then identify all valid keywords from \texttt{option\_list} appearing in the answer text to form the gold answer set \texttt{answer\_set}.

Similarly, in the prediction extraction step, we parse the model output \texttt{prediction} to build the predicted category set \texttt{prediction\_set}, based on the same keywords defined in \texttt{option\_list}.

We compare the predicted set with the gold set to compute the F1 score. The formula is defined as:
\begin{equation}
F_1 = \frac{2 \times \text{Precision} \times \text{Recall}}{\text{Precision} + \text{Recall}}
\end{equation}

Where:
\begin{equation}
\text{Precision} = \frac{TP}{TP + FP}, \quad \text{Recall} = \frac{TP}{TP + FN}
\end{equation}

\begin{itemize}
  \item \textbf{True Positives (TP)}: Number of predicted tags correctly matching the gold set ($\lvert \texttt{prediction\_set} \cap \texttt{answer\_set} \rvert$)
  \item \textbf{False Positives (FP)}: Predicted tags that do not belong to the gold set ($\lvert \texttt{prediction\_set} \setminus \texttt{answer\_set} \rvert$)
  \item \textbf{False Negatives (FN)}: Gold tags missed by the model ($\lvert \texttt{answer\_set} \setminus \texttt{prediction\_set} \rvert$)
\end{itemize}

We illustrate the evaluation process using one welfare compensation sample. The instruction reads: ``You are a legal expert in labor and social security. Based on the court decision, identify all relevant compensation categories in this case. Categories include: heat stroke allowance, high temperature subsidy, meal allowance, etc." 

\noindent The question is based on a labor dispute in which the plaintiff claimed multiple compensation types due to an illegal termination. The gold answer is: ``\texttt{Category} breach compensation <eoa>".

Suppose the model prediction is: ``\texttt{Category} holiday bonus, breach compensation, economic compensation <eoa>". We extract the legal labels from \texttt{option\_list}, resulting in a predicted tag set of: breach compensation, economic compensation. 


In this case:
\begin{itemize}[leftmargin=1.2em, itemsep=0pt, topsep=2pt]
    \item $\mathrm{TP}=1$ (\texttt{breach compensation})
    \item $\mathrm{FP}=2$ (\texttt{holiday bonus}, \texttt{economic compensation})
    \item $\mathrm{FN}=0$
\end{itemize}

Therefore:
\begin{equation}
\text{Precision} = \frac{1}{3}, \quad \text{Recall} = 1.0, \quad F_1 = \frac{2 \times \frac{1}{3} \times 1}{\frac{1}{3} + 1} = 0.5
\end{equation}

\subsection{Soft-\texorpdfstring{$F_1$}{F1} Evaluation for Named Entity Recognition}

We use a character-level Soft-$F_1$ to evaluate \textbf{T6} (named entity recognition). Unlike exact-match $F_1$ that requires token-identical spans, Soft-$F_1$ assigns partial credit when a predicted mention overlaps a gold mention, which is crucial in the legal domain where minor format variations (e.g., spacing, punctuation, or orthographic variants) should not erase substantial semantic correctness.

Instances instruct the model to extract entities in a structured line format, e.g.,
\texttt{[Entity Type] Entity Mention <eoa>}. We parse both the \texttt{answer} (gold) and \texttt{prediction} fields to obtain sets of typed mentions. To ensure comparability, we normalize each mention by (i) lowercasing, (ii) stripping punctuation and spaces (including full-/half-width variants), (iii) Unicode NFKC normalization, and (iv) representing each mention as a sequence of characters (for Chinese) or Unicode grapheme clusters (for non-Chinese text). Duplicates are removed. Mentions are evaluated within type (employee, employer, labor organization, etc.); mention order is ignored.

\textit{Pairing and per-mention scores.} For each entity type, let $G=\{g_1,\dots,g_m\}$ be gold mentions and $\widehat{G}=\{\hat g_1,\dots,\hat g_n\}$ be predicted mentions. We perform one-to-one pairing between $G$ and $\widehat{G}$ to maximize total character overlap (Hungarian matching on the bipartite graph with edge weight $|\hat g \cap g|$). For a matched pair $(\hat g, g)$, define
\begin{align}
TP(\hat g,g) &= |\hat g \cap g|, \quad
FP(\hat g,g) = |\hat g|-TP(\hat g,g), \\
FN(\hat g,g) &= |g|-TP(\hat g,g),
\end{align}
where $|\cdot|$ counts characters and $\cap$ is multiset overlap. Unmatched predictions are treated as pairs with an empty gold ($TP=0,\;FP=|\hat g|,\;FN=0$); unmatched gold mentions analogously have $TP=0,\;FP=0,\;FN=|g|$. For each matched (or unmatched) pair we compute
\begin{equation}
P=\frac{TP}{TP+FP},\qquad
R=\frac{TP}{TP+FN},\qquad
F_1=\frac{2PR}{P+R}.
\end{equation}

\textit{Aggregation.} The sample-level Soft-$F_1$ averages the per-pair $F_1$ over all pairs in the instance (including unmatched ones, which contribute $F_1=0$). At the corpus level, we report \emph{micro} Soft-$F_1$ by pooling character counts across all pairs and instances:
\begin{equation}
\begin{aligned}
\text{Micro-}P &= \frac{\sum TP}{\sum (TP+FP)},\quad
\text{Micro-}R = \frac{\sum TP}{\sum (TP+FN)},\\
\text{Micro-}F_1 &= \frac{2\,\text{Micro-}P\,\text{Micro-}R}{\text{Micro-}P+\text{Micro-}R}.
\end{aligned}
\end{equation}

We also report \emph{macro} Soft-$F_1$ as the unweighted mean of sample-level scores where noted.

\textit{Illustrative example.} Gold mentions:
\texttt{[Employee] Zhang San}, \texttt{[Employer] Hope Kindergarten}, \texttt{[Organization] Trade Union}.
Prediction:
\texttt{[Employee] Zhang San}, \texttt{[Employer] Hope Kindergarten}.
After normalization, the first two types are perfect matches ($F_1=1$ each); the organization is missing ($F_1=0$). The sample Soft-$F_1$ is
\[
\mathrm{Soft}\text{-}F_1^{\text{avg}}=\frac{1+1+0}{3}=0.6667.
\]
This illustrates that Soft-$F_1$ preserves full credit for exact matches while assigning zero credit to omissions, and in general provides graded credit for partial string overlaps.

\textit{Implementation notes.} (a) If multiple predictions overlap the same gold mention, only the best-matching prediction is paired; others are counted as unmatched predictions. (b) For nested or overlapping gold mentions within a type, pairing still enforces one-to-one matching. (c) If an instance legitimately contains no entities of a type, predictions of that type are penalized as unmatched predictions.

\section{Experiments}
\label{sec:experiment}
\begin{figure*}[!t]
 \caption{The Performance of different models.}
  \centering
  \includegraphics[width=\textwidth]{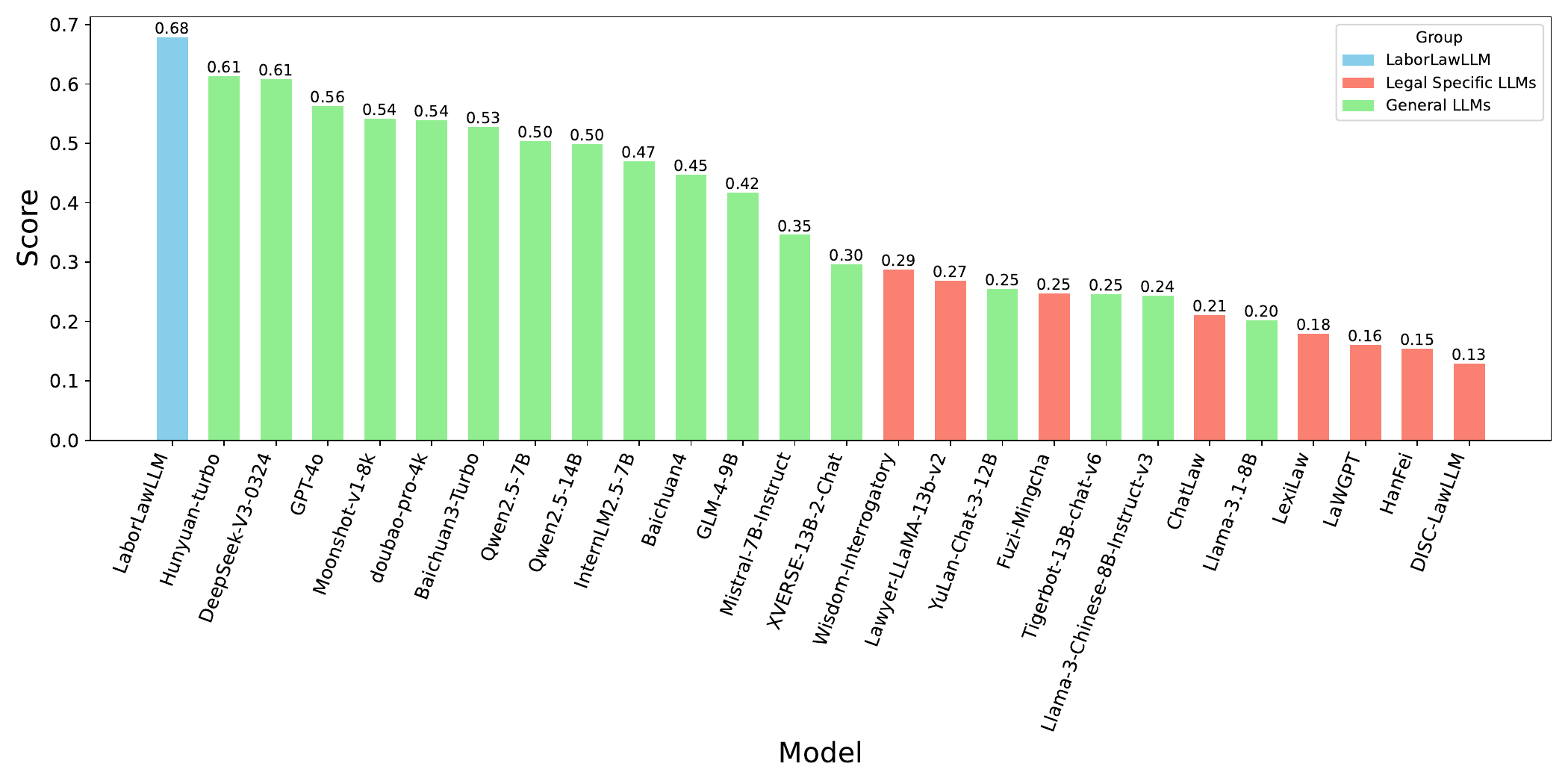}
  \label{fig:performance}
\end{figure*}

We present empirical results on the proposed LaborLaw benchmark. In total, we evaluate 26 systems—17 general-purpose LLMs (open and closed), 8 Chinese legal LLMs, and our domain-specialized \textbf{LaborLawLLM}—across 12 task types (T$_1$–T$_{12}$) and 12 case categories (C$_1$–C$_{12}$). An overview of all evaluated models, including their sizes, domains, and sources, is summarized in Table~\ref{tab:model_comparison}. For classification and extraction tasks we report Accuracy and F1/Soft-F1; for generation we use ROUGE-L; and for long-form judgments we adopt an LLM-as-judge protocol (GPT-o1). We visualize refusal behavior via cell shading, where darker cells denote higher abstention. We begin with overall performance and the aggregate leaderboard (Figure~\ref{fig:performance}), proceed to a task-level analysis (Table~\ref{tab:model_per_task}), and conclude with a stratification by case type (Table~\ref{tab:model_per_case}).

\begin{table*}[!t]
\centering
\caption{The comparisons of general and legal-domain language models.}
\label{tab:model_comparison}

\scriptsize
\setlength{\tabcolsep}{3pt}
\renewcommand{\arraystretch}{1.05}

\newcolumntype{L}{>{\RaggedRight\arraybackslash}X}
\newcolumntype{C}[1]{>{\centering\arraybackslash}p{#1}}

\begin{tabularx}{\linewidth}{%
L  C{0.9cm} C{1.1cm} C{1.0cm}  L  L  C{0.7cm}}
\toprule
\textbf{Model} & \textbf{Size} & \textbf{Domain} & \textbf{From} &
\textbf{Base model} & \textbf{Creator} & \textbf{Link} \\
\midrule
GPT-4o (\cite{openai2024gpt4})      & --  & General & API     & --              & OpenAI        & -- \\
LLaMA-3.1-8B (\cite{llama2023})     & 8B  & General & Weights & --              & Meta AI       & \href{https://github.com/meta-llama/llama3}{[1]} \\
Qwen2.5-7B (\cite{qwen2023})        & 7B  & General & Weights & --              & Alibaba       & \href{https://github.com/QwenLM/Qwen2.5}{[2]} \\
Qwen2.5-14B                         & 14B & General & Weights & --              & Alibaba       & \href{https://github.com/QwenLM/Qwen2.5}{[3]} \\
GLM-4-9B (\cite{chatglm2024})       & 9B  & General & Weights & --              & Zhipu AI      & \href{https://github.com/THUDM/GLM-4}{[4]} \\
DeepSeek-v3-0324 (\cite{deepseek2024}) & -- & General & API  & --              & DeepSeek AI   & \href{https://www.deepseek.com/}{[5]} \\
Moonshot                             & --  & General & API     & --              & Moonshot AI   & \href{https://platform.moonshot.cn/docs/intro}{[6]} \\
Baichuan4 (\cite{baichuan2023})      & --  & General & API     & --              & Baichuan Inc  & \href{https://platform.baichuan-ai.com/homePage}{[7]} \\
Baichuan3-Turbo                      & --  & General & API     & --              & Baichuan Inc  & \href{https://platform.baichuan-ai.com/homePage}{[8]} \\
Hunyuan (\cite{hunyuan2024})         & --  & General & API     & --              & Tencent       & \href{https://github.com/Tencent/Tencent-Hunyuan-Large}{[9]} \\
Mistral (\cite{mistral2023})         & 7B  & General & API     & --              & Mistral AI    & \href{https://github.com/mistralai/mistral-inference}{[10]} \\
Doubao                               & --  & General & API     & --              & ByteDance     & \href{https://www.doubao.com/chat/}{[11]} \\
InternLM2.5-7B (\cite{internlm2024}) & 7B  & General & Weights & --              & SenseTime     & \href{https://github.com/InternLM/InternLM}{[12]} \\
Tigerbot-13B-chat-v6 (\cite{tigerbot2023}) & 13B & General & Weights & --        & Tigerobo      & \href{https://github.com/TigerResearch/TigerBot}{[13]} \\
LLaMA-3-Chinese-8B-Instruct-v3       & 8B  & General & Weights & --              & YMCUI         & \href{https://github.com/ymcui/Chinese-LLaMA-Alpaca-3}{[14]} \\
XVERSE-13B-2-Chat                    & 13B & General & Weights & --              & Yuanxiang     & \href{https://github.com/xverse-ai/XVERSE-13B}{[15]} \\
YuLan-Chat-3-12B (\cite{yulan2024})  & 12B & General & Weights & --              & RUC-GSAI      & \href{https://github.com/RUC-GSAI/YuLan-Chat}{[16]} \\
HanFei (\cite{HanFei})              & 7B  & Law     & Weights & HanFei          & SIAT NLP      & \href{https://github.com/siat-nlp/HanFei}{[17]} \\
ChatLaw (\cite{chatlawmoe2023})      & 13B & Law     & Weights & Ziya-LLaMA-13B  & PKU-YuanGroup & \href{https://github.com/PKU-YuanGroup/ChatLaw}{[18]} \\
Lawyer-LLaMA-13b-v2 (\cite{lawyerllama2023}) & 13B & Law & Weights & LLaMA-Chinese-13B & Quzhe Huang & \href{https://github.com/AndrewZhe/lawyer-llama}{[19]} \\
LexiLaw                              & 6B  & Law     & Weights & ChatGLM-6B      & Haitao Li     & \href{https://github.com/CSHaitao/LexiLaw}{[20]} \\
LawGPT (\cite{lawgpt2024})           & 7B  & Law     & Weights & Chinese-LLaMA-7B& Pengxiao Song & \href{https://github.com/pengxiao-song/LaWGPT}{[21]} \\
Wisdom-Interrogatory                 & 7B  & Law     & Weights & Baichuan-7B     & ZJU/Alibaba   & \href{https://github.com/zhihaiLLM/wisdomInterrogatory}{[22]} \\
Fuzi-Mingcha (\cite{sdu_fuzi_mingcha}) & 6B & Law   & Weights & ChatGLM-6B      & irlab-sdu     & \href{https://github.com/irlab-sdu/fuzi.mingcha}{[23]} \\
DISC-LawLLM (\cite{disc2023})        & 13B & Law     & Weights & Baichuan-13B    & Fudan-DISC    & \href{https://github.com/FudanDISC/DISC-LawLLM}{[24]} \\
\bottomrule
\end{tabularx}
\end{table*}

\subsection{Overall Results on the LaborLaw Benchmark}
We evaluate 26 systems on our new LaborLaw benchmark: 17 general-purpose LLMs (open and closed), 8 Chinese legal LLMs, and our labor-law–specialized \textbf{LaborLawLLM}. We report a single aggregate score (higher is better) averaged over all benchmark tasks and visualize the leaderboard in Figure~\ref{fig:performance}.

\textbf{LaborLawLLM} ranks first with an overall score of 0.68, establishing the state of the art on this benchmark. The next strongest systems—DeepSeek-v3-0324 and Hunyuan-turbo—each reach 0.61, followed by GPT-4o at 0.56. Beyond headline accuracy, the margin between our model and the closest competitors is consistent across task groups and case categories (Tables~\ref{tab:model_per_task} and \ref{tab:model_per_case}), indicating that the gains are not driven by a single subtask but by broad improvements in labor-law understanding.

Comparisons to closely related backbones highlight the contribution of domain adaptation. Untuned Qwen2.5-7B and Qwen2.5-14B baselines both score around 0.50 (Tables~\ref{tab:model_per_task} and \ref{tab:model_per_case}). With comparable parameter scale, our labor-law–trained variant delivers a +0.18 absolute gain, pointing to targeted data and alignment—rather than model size—as the primary driver.

General-purpose Chinese LLMs mostly cluster between 0.30 and 0.55, while prior legal-domain models lag substantially: the strongest legal baseline peaks at about 0.29 (e.g., Lawyer-LLaMA-13B-v2), and others—including YuLan-Chat-3-12B, Fuzi-Mingcha, Tigerbot-13B-chat-v6, Llama-3-Chinese-8B-Instruct-v3, ChatLaw, LaWGPT, LexiLaw, HanFei, and DISC-LawLLM—fall in the 0.13–0.27 range.

\subsection{Task-Level Results on the LaborLaw Benchmark}

\begin{table*}[!t]
\centering
\caption{The Performance of Baselines on LaborLaw Benchmark across Different Tasks. 
Note: Metric abbreviations --- R--L: ROUGE-L, Acc: Accuracy, F1: F1 Score, S--F1: Soft-F1, G--O1: GPT-o1. 
Color legend (\%abstention): \colorbox{white}{\strut 0\%}\;
\colorbox{mygreen!25}{\strut 25\%}\;
\colorbox{mygreen!50}{\strut 50\%}\;
\colorbox{mygreen!75}{\strut 75\%}\;
\colorbox{mygreen!100}{\strut 100\%}.}
\label{tab:model_per_task}

\begin{threeparttable}
\renewcommand{\arraystretch}{1.15}
\setlength{\tabcolsep}{5pt}

\resizebox{\linewidth}{!}{%
\begin{tabular}{lccccccccccccc}
\toprule
\multirow{2}{*}{Model}
& $T_1$ & $T_2$ & $T_3$ & $T_4$ & $T_5$ & $T_6$ & $T_7$ & $T_8$ & $T_9$ & $T_{10}$ & $T_{11}$ & $T_{12}$ & \multirow{2}{*}{Avg} \\
& R\text{-}L & Acc & Acc & Acc & F1 & S\text{-}F1 & G\text{-}O1 & G\text{-}O1 & G\text{-}O1 & R\text{-}L & G\text{-}O1 & G\text{-}O1 \\
\midrule

LaborLawLLM & \cellcolor{white}0.56 & \cellcolor{white}1 & \cellcolor{white}0.95 & \cellcolor{white}0.99 & \cellcolor{white}0.96 & \cellcolor{white}0.98 & \cellcolor{white}0.36 & \cellcolor{white}0.57 & \cellcolor{white}0.43 & \cellcolor{white}0.11 & \cellcolor{white}0.59 & \cellcolor{white}0.65 & 0.68 \\
\midrule

DeepSeek\text{-}v3\text{-}0324 & \cellcolor{white}0.42 & \cellcolor{white}0.84 & \cellcolor{white}0.64 & \cellcolor{white}0.75 & \cellcolor{white}0.62 & \cellcolor{white}0.79 & \cellcolor{white}0.50 & \cellcolor{white}0.69 & \cellcolor{white}0.35 & \cellcolor{white}0.34 & \cellcolor{white}0.66 & \cellcolor{white}0.70 & 0.61 \\
Hunyuan\text{-}turbo & \cellcolor{white}0.66 & \cellcolor{white}0.88 & \cellcolor{white}0.86 & \cellcolor{white}0.71 & \cellcolor{white}0.60 & \cellcolor{white}0.76 & \cellcolor{white}0.46 & \cellcolor{white}0.62 & \cellcolor{white}0.30 & \cellcolor{white}0.28 & \cellcolor{white}0.59 & \cellcolor{white}0.65 & 0.61 \\
GPT\text{-}4o & \cellcolor{white}0.17 & \cellcolor{white}0.80 & \cellcolor{white}0.49 & \cellcolor{white}0.74 & \cellcolor{white}0.60 & \cellcolor{white}0.84 & \cellcolor{white}0.49 & \cellcolor{white}0.70 & \cellcolor{white}0.34 & \cellcolor{white}0.26 & \cellcolor{white}0.65 & \cellcolor{white}0.70 & 0.57 \\
doubao\text{-}pro\text{-}4k & \cellcolor{white}0.41 & \cellcolor{white}0.81 & \cellcolor{white}0.58 & \cellcolor{white}0.65 & \cellcolor{mygreen!25}0.37 & \cellcolor{white}0.86 & \cellcolor{white}0.45 & \cellcolor{white}0.59 & \cellcolor{white}0.29 & \cellcolor{white}0.36 & \cellcolor{white}0.54 & \cellcolor{white}0.56 & 0.54 \\
Moonshot\text{-}v1\text{-}8k & \cellcolor{white}0.35 & \cellcolor{white}0.72 & \cellcolor{white}0.47 & \cellcolor{white}0.71 & \cellcolor{white}0.45 & \cellcolor{white}0.86 & \cellcolor{white}0.46 & \cellcolor{white}0.63 & \cellcolor{white}0.33 & \cellcolor{white}0.30 & \cellcolor{white}0.59 & \cellcolor{white}0.62 & 0.54 \\
Baichuan3\text{-}Turbo & \cellcolor{mygreen!50}0.12 & \cellcolor{white}0.75 & \cellcolor{white}0.60 & \cellcolor{white}0.73 & \cellcolor{white}0.49 & \cellcolor{white}0.82 & \cellcolor{white}0.47 & \cellcolor{white}0.67 & \cellcolor{white}0.30 & \cellcolor{white}0.22 & \cellcolor{white}0.56 & \cellcolor{white}0.61 & 0.53 \\
Qwen2.5\text{-}7B & \cellcolor{white}0.19 & \cellcolor{white}0.70 & \cellcolor{white}0.48 & \cellcolor{white}0.65 & \cellcolor{white}0.48 & \cellcolor{white}0.81 & \cellcolor{white}0.46 & \cellcolor{white}0.64 & \cellcolor{white}0.29 & \cellcolor{white}0.22 & \cellcolor{white}0.56 & \cellcolor{white}0.58 & 0.51 \\
Qwen2.5\text{-}14B & \cellcolor{white}0.11 & \cellcolor{white}0.78 & \cellcolor{white}0.42 & \cellcolor{white}0.74 & \cellcolor{white}0.50 & \cellcolor{white}0.87 & \cellcolor{white}0.47 & \cellcolor{white}0.61 & \cellcolor{white}0.35 & \cellcolor{white}0.07 & \cellcolor{white}0.51 & \cellcolor{white}0.58 & 0.50 \\
InternLM2.5\text{-}7B & \cellcolor{white}0.17 & \cellcolor{white}0.75 & \cellcolor{white}0.43 & \cellcolor{white}0.56 & \cellcolor{white}0.43 & \cellcolor{white}0.61 & \cellcolor{white}0.44 & \cellcolor{white}0.66 & \cellcolor{white}0.31 & \cellcolor{white}0.19 & \cellcolor{white}0.53 & \cellcolor{white}0.55 & 0.47 \\
Baichuan4 & \cellcolor{white}0.22 & \cellcolor{white}0.65 & \cellcolor{white}0.40 & \cellcolor{white}0.66 & \cellcolor{white}0.25 & \cellcolor{white}0.52 & \cellcolor{white}0.46 & \cellcolor{white}0.67 & \cellcolor{white}0.28 & \cellcolor{mygreen!50}0.08 & \cellcolor{white}0.56 & \cellcolor{white}0.60 & 0.45 \\
GLM\text{-}4\text{-}9B & \cellcolor{white}0.12 & \cellcolor{white}0.63 & \cellcolor{white}0.40 & \cellcolor{white}0.35 & \cellcolor{white}0.29 & \cellcolor{white}0.76 & \cellcolor{white}0.40 & \cellcolor{white}0.58 & \cellcolor{white}0.26 & \cellcolor{white}0.18 & \cellcolor{white}0.49 & \cellcolor{white}0.55 & 0.42 \\
Mistral & \cellcolor{white}0.09 & \cellcolor{white}0.24 & \cellcolor{white}0.04 & \cellcolor{white}0.47 & \cellcolor{white}0.50 & \cellcolor{white}0.71 & \cellcolor{white}0.38 & \cellcolor{white}0.53 & \cellcolor{white}0.22 & \cellcolor{white}0.16 & \cellcolor{white}0.41 & \cellcolor{white}0.42 & 0.35 \\
XVERSE\text{-}13B\text{-}2\text{-}Chat & \cellcolor{white}0.10 & \cellcolor{white}0.27 & \cellcolor{white}0.14 & \cellcolor{white}0.29 & \cellcolor{white}0.27 & \cellcolor{white}0.45 & \cellcolor{white}0.36 & \cellcolor{white}0.52 & \cellcolor{white}0.23 & \cellcolor{white}0.17 & \cellcolor{white}0.38 & \cellcolor{white}0.38 & 0.30 \\
YuLan\text{-}Chat\text{-}3\text{-}12B & \cellcolor{white}0.13 & \cellcolor{white}0.23 & \cellcolor{white}0.02 & \cellcolor{white}0.08 & \cellcolor{white}0.19 & \cellcolor{white}0.35 & \cellcolor{white}0.37 & \cellcolor{white}0.52 & \cellcolor{white}0.26 & \cellcolor{white}0.17 & \cellcolor{white}0.38 & \cellcolor{white}0.35 & 0.25 \\
Tigerbot\text{-}13B\text{-}chat\text{-}v6 & \cellcolor{white}0.15 & \cellcolor{white}0.01 & \cellcolor{white}0.06 & \cellcolor{white}0.09 & \cellcolor{mygreen!25}0.06 & \cellcolor{white}0.34 & \cellcolor{white}0.37 & \cellcolor{white}0.47 & \cellcolor{mygreen!50}0.28 & \cellcolor{white}0.21 & \cellcolor{mygreen!25}0.45 & \cellcolor{mygreen!25}0.47 & 0.25 \\
Llama\text{-}3\text{-}Chinese\text{-}8B\text{-}Instruct\text{-}v3 & \cellcolor{white}0.05 & \cellcolor{mygreen!25}0.13 & \cellcolor{mygreen!25}0.10 & \cellcolor{white}0.13 & \cellcolor{white}0.34 & \cellcolor{white}0.66 & \cellcolor{white}0.26 & \cellcolor{white}0.27 & \cellcolor{white}0.21 & \cellcolor{white}0.14 & \cellcolor{white}0.33 & \cellcolor{white}0.31 & 0.24 \\
Llama\text{-}3.1\text{-}8B & \cellcolor{mygreen!50}0.03 & \cellcolor{mygreen!75}0.09 & \cellcolor{mygreen!75}0.01 & \cellcolor{white}0.05 & \cellcolor{mygreen!50}0.14 & \cellcolor{white}0.46 & \cellcolor{mygreen!25}0.28 & \cellcolor{mygreen!25}0.38 & \cellcolor{mygreen!50}0.23 & \cellcolor{mygreen!25}0.08 & \cellcolor{mygreen!25}0.35 & \cellcolor{mygreen!50}0.35 & 0.20 \\
\midrule
Wisdom\text{-}Interrogatory & \cellcolor{white}0.26 & \cellcolor{white}0.17 & \cellcolor{white}0.03 & \cellcolor{white}0.08 & \cellcolor{mygreen!20}0.22 & \cellcolor{white}0.35 & \cellcolor{white}0.41 & \cellcolor{white}0.52 & \cellcolor{white}0.25 & \cellcolor{white}0.23 & \cellcolor{white}0.45 & \cellcolor{white}0.48 & 0.29 \\
Lawyer\text{-}LLaMA\text{-}13b\text{-}v2 & \cellcolor{white}0.13 & \cellcolor{white}0.34 & \cellcolor{white}0.01 & \cellcolor{white}0.05 & \cellcolor{mygreen!25}0.33 & \cellcolor{white}0.38 & \cellcolor{white}0.39 & \cellcolor{white}0.45 & \cellcolor{white}0.26 & \cellcolor{white}0.14 & \cellcolor{white}0.36 & \cellcolor{white}0.41 & 0.27 \\
Fuzi\text{-}Mingcha & \cellcolor{white}0.15 & \cellcolor{white}0.11 & \cellcolor{white}0.19 & \cellcolor{white}0.11 & \cellcolor{mygreen!25}0.18 & \cellcolor{white}0.33 & \cellcolor{white}0.39 & \cellcolor{white}0.44 & \cellcolor{white}0.24 & \cellcolor{white}0.15 & \cellcolor{white}0.39 & \cellcolor{white}0.30 & 0.25 \\
ChatLaw & \cellcolor{white}0.10 & \cellcolor{mygreen!25}0.15 & \cellcolor{mygreen!25}0.02 & \cellcolor{white}0.07 & \cellcolor{mygreen!25}0.19 & \cellcolor{white}0.34 & \cellcolor{white}0.33 & \cellcolor{white}0.38 & \cellcolor{white}0.21 & \cellcolor{white}0.18 & \cellcolor{white}0.29 & \cellcolor{white}0.29 & 0.21 \\
LexiLaw & \cellcolor{white}0.09 & \cellcolor{white}0.24 & \cellcolor{white}0.11 & \cellcolor{white}0.00 & \cellcolor{white}0.09 & \cellcolor{white}0.14 & \cellcolor{white}0.27 & \cellcolor{white}0.34 & \cellcolor{white}0.24 & \cellcolor{white}0.13 & \cellcolor{white}0.25 & \cellcolor{white}0.27 & 0.18 \\
LaWGPT & \cellcolor{white}0.01 & \cellcolor{white}0.24 & \cellcolor{white}0.10 & \cellcolor{white}0.00 & \cellcolor{mygreen!25}0.12 & \cellcolor{white}0.18 & \cellcolor{white}0.29 & \cellcolor{white}0.24 & \cellcolor{white}0.23 & \cellcolor{white}0.07 & \cellcolor{white}0.22 & \cellcolor{white}0.23 & 0.16 \\
HanFei & \cellcolor{white}0.17 & \cellcolor{mygreen!40}0.08 & \cellcolor{mygreen!50}0.02 & \cellcolor{white}0.00 & \cellcolor{mygreen!50}0.00 & \cellcolor{white}0.38 & \cellcolor{white}0.22 & \cellcolor{white}0.20 & \cellcolor{white}0.20 & \cellcolor{white}0.18 & \cellcolor{white}0.20 & \cellcolor{white}0.20 & 0.15 \\
DISC\text{-}LawLLM & \cellcolor{white}0.14 & \cellcolor{mygreen!25}0.00 & \cellcolor{white}0.00 & \cellcolor{white}0.00 & \cellcolor{mygreen!75}0.00 & \cellcolor{white}0.38 & \cellcolor{white}0.20 & \cellcolor{white}0.20 & \cellcolor{white}0.20 & \cellcolor{white}0.03 & \cellcolor{white}0.20 & \cellcolor{white}0.20 & 0.13 \\
\bottomrule
\end{tabular}%
}

\end{threeparttable}
\end{table*}

On the task-level benchmark in Table~\ref{tab:model_per_task}, the purpose-built \textbf{LaborLawLLM} attains the highest average across all twelve tasks (0.68), clearly outperforming the strongest general-purpose models (0.61) and far exceeding the best prior legal-domain baseline (0.29). Visual shading in the table encodes refusal behavior; darker cells indicate higher abstention. \textbf{LaborLawLLM} exhibits consistently light shading, suggesting that domain training improves both accuracy and willingness to answer legally framed queries.
A compact radar view of the per-task pattern (T$_1$–T$_{12}$) is shown in Figure~\ref{fig:rader_renwu}.

By task, the benefits of specialization are most pronounced on knowledge-centric and classification problems. \textbf{LaborLawLLM} reaches 1.00 on single-choice knowledge QA (T2), 0.95 on multi-choice (T3), and 0.99 on cause-of-action prediction (T4). It also delivers strong F1 on welfare compensation (T5, 0.96) and named-entity recognition (T6, 0.98), indicating a high ceiling for structured decision tasks. For information-mining tasks evaluated by an LLM-as-judge (GPT-o1), results are mixed: the model trails leading general systems on claim mining (T7, 0.36) and dispute-focus extraction (T8, 0.57), but surpasses them on fact-based provision prediction (T9, 0.43), with an absolute gain of approximately 0.08–0.13 over strong general baselines. Across these extraction tasks, many smaller baselines show substantial refusal rates (25--75\%), whereas our model remains more willing to engage. On scenario-based provision prediction (T10), \textbf{LaborLawLLM}'s ROUGE-L (0.11) lags frontier general models; manual inspection suggests a formatting mismatch—concise statutory citations versus longer rationales rewarded by ROUGE—highlighting opportunities in instruction design, decoding, and training-signal shaping rather than gaps in legal knowledge. For long-form case analysis (T11–T12), also judged by GPT-o1, the model is competitive (0.59 and 0.65, respectively) while maintaining low abstention.

\subsection[Case-Type Results on the LaborLaw Benchmark (C1--C12)]%
{Case-Type Results on the LaborLaw Benchmark (C$_1$--C$_{12}$)}

\begin{table*}[!t]
\centering
\caption{The Performance of Baselines on LaborLaw Benchmark across case types. $C_1$–$C_{12}$ correspond to the case types listed in Table~\ref{tab:c1c12_vertical}. Color legend (\%abstention): \colorbox{white}{\strut 0\%}\; \colorbox{mygreen!25}{\strut 25\%}\; \colorbox{mygreen!50}{\strut 50\%}\; \colorbox{mygreen!75}{\strut 75\%}\; \colorbox{mygreen!100}{\strut 100\%}.
}
\label{tab:model_per_case}
\begin{threeparttable}
\renewcommand{\arraystretch}{1.15}
\setlength{\tabcolsep}{5pt}

\resizebox{\linewidth}{!}{%
\begin{tabular}{lccccccccccccc}
\toprule
Model  & $C_1$ & $C_2$ & $C_3$ & $C_4$ & $C_5$ & $C_6$ & $C_7$ & $C_8$ & $C_9$ & $C_{10}$ & $C_{11}$ & $C_{12}$ & Avg \\
\midrule
LaborLawLLM & \cellcolor{white}0.66 & \cellcolor{white}0.67 & \cellcolor{white}0.67 & \cellcolor{white}0.65 & \cellcolor{white}0.68 & \cellcolor{white}0.67 & \cellcolor{white}0.67 & \cellcolor{white}0.72 & \cellcolor{white}0.68 & \cellcolor{white}0.67 & \cellcolor{white}0.69 & \cellcolor{white}0.70 & 0.68 \\
\midrule
DeepSeek-v3     & \cellcolor{white}0.60 & \cellcolor{white}0.59 & \cellcolor{white}0.62 & \cellcolor{white}0.56 & \cellcolor{white}0.61 & \cellcolor{white}0.64 & \cellcolor{white}0.61 & \cellcolor{white}0.62 & \cellcolor{white}0.60 & \cellcolor{white}0.62 & \cellcolor{white}0.61 & \cellcolor{white}0.62 & 0.61 \\
hunyuan-turbo   & \cellcolor{white}0.60 & \cellcolor{white}0.57 & \cellcolor{white}0.63 & \cellcolor{white}0.56 & \cellcolor{white}0.59 & \cellcolor{white}0.65 & \cellcolor{white}0.62 & \cellcolor{white}0.64 & \cellcolor{white}0.61 & \cellcolor{white}0.65 & \cellcolor{white}0.63 & \cellcolor{white}0.63 & 0.61 \\
GPT-4o          & \cellcolor{white}0.55 & \cellcolor{white}0.53 & \cellcolor{white}0.58 & \cellcolor{white}0.52 & \cellcolor{white}0.57 & \cellcolor{white}0.59 & \cellcolor{white}0.55 & \cellcolor{white}0.60 & \cellcolor{white}0.56 & \cellcolor{white}0.56 & \cellcolor{white}0.57 & \cellcolor{white}0.58 & 0.56 \\
doubao-pro-4k   & \cellcolor{white}0.54 & \cellcolor{white}0.51 & \cellcolor{white}0.56 & \cellcolor{white}0.45 & \cellcolor{white}0.53 & \cellcolor{white}0.54 & \cellcolor{white}0.57 & \cellcolor{white}0.58 & \cellcolor{white}0.54 & \cellcolor{white}0.54 & \cellcolor{white}0.56 & \cellcolor{white}0.56 & 0.54 \\
Moonshot-v1-8k  & \cellcolor{white}0.53 & \cellcolor{white}0.51 & \cellcolor{white}0.57 & \cellcolor{white}0.48 & \cellcolor{white}0.53 & \cellcolor{white}0.57 & \cellcolor{white}0.55 & \cellcolor{white}0.53 & \cellcolor{white}0.52 & \cellcolor{white}0.59 & \cellcolor{white}0.57 & \cellcolor{white}0.54 & 0.54 \\
Baichuan3-Turbo & \cellcolor{white}0.52 & \cellcolor{white}0.50 & \cellcolor{white}0.54 & \cellcolor{white}0.49 & \cellcolor{white}0.53 & \cellcolor{white}0.53 & \cellcolor{white}0.55 & \cellcolor{white}0.53 & \cellcolor{white}0.51 & \cellcolor{white}0.55 & \cellcolor{white}0.54 & \cellcolor{white}0.54 & 0.53 \\
Qwen2.5-14B     & \cellcolor{white}0.46 & \cellcolor{white}0.45 & \cellcolor{white}0.50 & \cellcolor{white}0.49 & \cellcolor{white}0.50 & \cellcolor{white}0.53 & \cellcolor{white}0.52 & \cellcolor{white}0.53 & \cellcolor{white}0.47 & \cellcolor{white}0.53 & \cellcolor{white}0.50 & \cellcolor{white}0.51 & 0.50 \\
Qwen2.5-7B      & \cellcolor{white}0.49 & \cellcolor{white}0.47 & \cellcolor{white}0.50 & \cellcolor{white}0.46 & \cellcolor{white}0.50 & \cellcolor{white}0.54 & \cellcolor{white}0.51 & \cellcolor{white}0.51 & \cellcolor{white}0.46 & \cellcolor{white}0.54 & \cellcolor{white}0.53 & \cellcolor{white}0.53 & 0.50 \\
InternLM2.5-7B  & \cellcolor{white}0.44 & \cellcolor{white}0.44 & \cellcolor{white}0.48 & \cellcolor{white}0.42 & \cellcolor{white}0.45 & \cellcolor{white}0.51 & \cellcolor{white}0.47 & \cellcolor{white}0.51 & \cellcolor{white}0.46 & \cellcolor{white}0.50 & \cellcolor{white}0.47 & \cellcolor{white}0.49 & 0.47 \\
Baichuan4       & \cellcolor{white}0.40 & \cellcolor{white}0.41 & \cellcolor{white}0.48 & \cellcolor{white}0.42 & \cellcolor{white}0.44 & \cellcolor{white}0.49 & \cellcolor{white}0.46 & \cellcolor{white}0.43 & \cellcolor{white}0.44 & \cellcolor{white}0.48 & \cellcolor{white}0.44 & \cellcolor{white}0.48 & 0.44 \\
GLM-4-9B        & \cellcolor{white}0.39 & \cellcolor{white}0.39 & \cellcolor{white}0.41 & \cellcolor{white}0.41 & \cellcolor{white}0.39 & \cellcolor{white}0.45 & \cellcolor{white}0.45 & \cellcolor{white}0.41 & \cellcolor{white}0.41 & \cellcolor{white}0.44 & \cellcolor{white}0.41 & \cellcolor{white}0.44 & 0.42 \\
Mistral         & \cellcolor{white}0.32 & \cellcolor{white}0.31 & \cellcolor{white}0.37 & \cellcolor{white}0.31 & \cellcolor{white}0.35 & \cellcolor{white}0.39 & \cellcolor{white}0.37 & \cellcolor{white}0.35 & \cellcolor{white}0.33 & \cellcolor{white}0.38 & \cellcolor{white}0.32 & \cellcolor{white}0.36 & 0.35 \\
XVERSE-13B-2-Chat & \cellcolor{white}0.29 & \cellcolor{white}0.28 & \cellcolor{white}0.31 & \cellcolor{white}0.28 & \cellcolor{white}0.29 & \cellcolor{white}0.30 & \cellcolor{white}0.31 & \cellcolor{white}0.33 & \cellcolor{white}0.29 & \cellcolor{white}0.28 & \cellcolor{white}0.29 & \cellcolor{white}0.30 & 0.30 \\
YuLan-Chat-3-12B & \cellcolor{white}0.23 & \cellcolor{white}0.24 & \cellcolor{white}0.25 & \cellcolor{white}0.27 & \cellcolor{white}0.26 & \cellcolor{white}0.25 & \cellcolor{white}0.27 & \cellcolor{white}0.31 & \cellcolor{white}0.24 & \cellcolor{white}0.24 & \cellcolor{white}0.24 & \cellcolor{white}0.26 & 0.25 \\
Tigerbot-13B-chat-v6 & \cellcolor{mygreen!25}0.21 & \cellcolor{white}0.23 & \cellcolor{white}0.23 & \cellcolor{white}0.26 & \cellcolor{white}0.24 & \cellcolor{white}0.26 & \cellcolor{white}0.23 & \cellcolor{white}0.30 & \cellcolor{white}0.24 & \cellcolor{white}0.25 & \cellcolor{white}0.24 & \cellcolor{white}0.27 & 0.25 \\
Llama-3-Chinese-8B-Instruct-v3 & \cellcolor{white}0.26 & \cellcolor{white}0.24 & \cellcolor{white}0.23 & \cellcolor{white}0.21 & \cellcolor{white}0.24 & \cellcolor{white}0.27 & \cellcolor{white}0.26 & \cellcolor{white}0.24 & \cellcolor{white}0.25 & \cellcolor{white}0.25 & \cellcolor{white}0.24 & \cellcolor{white}0.26 & 0.24 \\
Llama-3.1-8B    & \cellcolor{mygreen!25}0.21 & \cellcolor{mygreen!25}0.19 & \cellcolor{mygreen!25}0.17 & \cellcolor{mygreen!25}0.21 & \cellcolor{mygreen!25}0.20 & \cellcolor{mygreen!25}0.19 & \cellcolor{mygreen!25}0.22 & \cellcolor{mygreen!25}0.22 & \cellcolor{mygreen!25}0.20 & \cellcolor{mygreen!50}0.21 & \cellcolor{mygreen!25}0.20 & \cellcolor{mygreen!25}0.19 & 0.20 \\
\midrule
Wisdom-Interrogatory & \cellcolor{white}0.27 & \cellcolor{white}0.26 & \cellcolor{white}0.28 & \cellcolor{white}0.29 & \cellcolor{white}0.29 & \cellcolor{white}0.27 & \cellcolor{white}0.28 & \cellcolor{white}0.37 & \cellcolor{white}0.29 & \cellcolor{white}0.28 & \cellcolor{white}0.28 & \cellcolor{white}0.29 & 0.29 \\
Lawyer-LLaMA-13b-v2 & \cellcolor{white}0.26 & \cellcolor{white}0.25 & \cellcolor{white}0.26 & \cellcolor{white}0.26 & \cellcolor{white}0.27 & \cellcolor{white}0.26 & \cellcolor{white}0.29 & \cellcolor{white}0.31 & \cellcolor{white}0.27 & \cellcolor{white}0.26 & \cellcolor{white}0.26 & \cellcolor{white}0.28 & 0.27 \\
Fuzi-Mingcha    & \cellcolor{white}0.23 & \cellcolor{white}0.22 & \cellcolor{white}0.23 & \cellcolor{white}0.25 & \cellcolor{white}0.22 & \cellcolor{white}0.24 & \cellcolor{white}0.26 & \cellcolor{white}0.34 & \cellcolor{white}0.23 & \cellcolor{white}0.26 & \cellcolor{white}0.25 & \cellcolor{white}0.24 & 0.25 \\
ChatLaw         & \cellcolor{white}0.20 & \cellcolor{white}0.20 & \cellcolor{white}0.21 & \cellcolor{white}0.22 & \cellcolor{white}0.20 & \cellcolor{white}0.21 & \cellcolor{white}0.24 & \cellcolor{white}0.24 & \cellcolor{white}0.21 & \cellcolor{white}0.20 & \cellcolor{white}0.20 & \cellcolor{white}0.20 & 0.21 \\
LexiLaw         & \cellcolor{white}0.18 & \cellcolor{white}0.17 & \cellcolor{white}0.20 & \cellcolor{white}0.18 & \cellcolor{white}0.17 & \cellcolor{white}0.18 & \cellcolor{white}0.18 & \cellcolor{white}0.20 & \cellcolor{white}0.18 & \cellcolor{white}0.18 & \cellcolor{white}0.17 & \cellcolor{white}0.17 & 0.18 \\
LaWGPT          & \cellcolor{white}0.16 & \cellcolor{white}0.15 & \cellcolor{white}0.17 & \cellcolor{white}0.17 & \cellcolor{white}0.17 & \cellcolor{white}0.16 & \cellcolor{white}0.15 & \cellcolor{white}0.16 & \cellcolor{white}0.16 & \cellcolor{white}0.16 & \cellcolor{white}0.16 & \cellcolor{white}0.15 & 0.16 \\
HanFei          & \cellcolor{white}0.15 & \cellcolor{white}0.15 & \cellcolor{white}0.14 & \cellcolor{white}0.18 & \cellcolor{white}0.16 & \cellcolor{white}0.15 & \cellcolor{white}0.16 & \cellcolor{white}0.15 & \cellcolor{white}0.16 & \cellcolor{white}0.15 & \cellcolor{white}0.15 & \cellcolor{white}0.16 & 0.15 \\
DISC-LawLLM     & \cellcolor{white}0.12 & \cellcolor{white}0.12 & \cellcolor{white}0.16 & \cellcolor{white}0.16 & \cellcolor{white}0.13 & \cellcolor{white}0.12 & \cellcolor{white}0.13 & \cellcolor{white}0.13 & \cellcolor{white}0.13 & \cellcolor{white}0.12 & \cellcolor{white}0.12 & \cellcolor{white}0.13 & 0.13 \\
\bottomrule
\end{tabular}%
}

\end{threeparttable}
\end{table*}

We stratify the evaluation by the twelve case types defined in Table~\ref{tab:c1c12_vertical} and report per-category results in Table~\ref{tab:model_per_case}. Across all twelve categories, the domain-trained \textbf{LaborLawLLM} attains the top average (0.68); the strongest general-purpose systems cluster around 0.61, while prior legal-domain baselines trail near 0.29. \textbf{LaborLawLLM} is strong and stable across categories (0.65--0.72), notably reaching 0.72 on C$_8$ (confirmation of labor relationship disputes) and 0.70 on C$_{12}$ (wage claims), and remaining above 0.66 on core categories such as C$_1$--C$_7$ (personnel and social-insurance–related disputes). This uniformity contrasts with the wider spread for general models, which often dip on categories requiring precise statutory triggers or specialized remedies (e.g., non-compete and severance compensation).

Cell shading in Table~\ref{tab:model_per_case} encodes refusal behavior; darker cells indicate higher abstention. \textbf{LaborLawLLM} shows consistently light shading across all C$_i$, reflecting low refusal even on categories that commonly induce safety-style abstentions in generic LLMs. Several smaller baselines display markedly darker cells in multiple C$_i$, indicating a tendency to withhold answers when prompts use domain-specific phrasing or require statute-grounded reasoning.

\section{Conclusion}
This paper introduces the \textit{LaborLaw} benchmark, a comprehensive evaluation suite for Chinese labor-law reasoning that spans 12 task types (T$_1$–T$_{12}$) and 12 case categories (C$_1$–C$_{12}$). Across 26 evaluated systems—17 general-purpose LLMs and 8 legal-domain LLMs—our domain-specialized \textbf{LaborLawLLM} establishes a new state of the art with an aggregate score of 0.68, outperforming the strongest general-purpose models (0.61) and substantially exceeding prior legal baselines (around 0.29). Beyond overall accuracy, \textbf{LaborLawLLM} exhibits consistently low refusal rates as visualized by cell shading, indicating that targeted domain training improves both correctness and willingness to engage with legally framed prompts.

At the task level, specialization yields especially strong gains on knowledge-centric and structured decisions (e.g., T2–T6), while long-form, judge-based mining tasks (T7–T9) reveal remaining headroom: the model underperforms on claim mining and dispute-focus extraction but surpasses general systems on fact-based provision prediction. Scenario-based provision prediction (T10) highlights a formatting sensitivity—concise statutory citations can be penalized under ROUGE—suggesting that instruction style and decoding choices materially affect measured performance. By case type, results are uniformly strong (0.65–0.72), peaking on confirmation of labor relationships (C$_8$) and wage claims (C$_{12}$), which require precise statute application and remedy calculation.

Looking forward, we see three priorities. \textit{(i) Data and alignment:} expand and refine labor-law supervision to strengthen judge-based mining tasks and reduce evaluation-format mismatch. \textit{(ii) Methodology:} improve instruction design and decoding strategies for open-ended generation while preserving statutory precision. \textit{(iii) Scope and robustness:} broaden coverage to additional subdomains and jurisdictions, and assess temporal robustness as statutes evolve. Together, these directions aim to further close gaps with frontier general models where they remain competitive, while preserving the clear advantages conferred by domain-aware training on the \textit{LaborLaw} benchmark.


\begin{thebibliography}{99}
\bibliographystyle{IEEEtran}

\bibitem{openai2024gpt4}
OpenAI et al., ``GPT-4 Technical Report,'' \textit{arXiv}, arXiv:2303.08774, 2024, ver. 6 (Mar. 4, 2024). [Online]. Available: \url{https://arxiv.org/abs/2303.08774}

\bibitem{chatlawmoe2023}
PKU YuanGroup, ``ChatLaw-MOE: Enhancing Legal Large Language Models with Mixture of Experts,'' PKU Legal AI Research Group, 2023. [Online]. Available: \url{https://github.com/PKUYuanGroup/ChatLaw}

\bibitem{disc2023}
Fudan University DISC Team, ``DISC-LawLLM: A Legal Domain-Specific Large Language Model,'' Fudan DISC AI Research, 2023. [Online]. Available: \url{https://github.com/FudanDISC/DISC-LawLLM}

\bibitem{lawgpt2024}
Z.~Zhou, J.-X.~Shi, P.-X.~Song, X.-W.~Yang, Y.-X.~Jin, L.-Z.~Guo, and Y.-F.~Li, ``LaWGPT: A Chinese Legal Knowledge-Enhanced Large Language Model,'' \textit{arXiv preprint} arXiv:2305.12345, 2024. [Online]. Available: \url{https://arxiv.org/abs/2305.12345}

\bibitem{lawyerllama2023}
Q.~Huang, M.~Tao, C.~Zhang, Z.~An, C.~Jiang, Z.~Chen, Z.~Wu, and Y.~Feng, ``Lawyer LLaMA: Enhancing LLMs with Legal Knowledge,'' \textit{arXiv preprint} arXiv:2305.15062, 2023. [Online]. Available: \url{https://arxiv.org/abs/2305.15062}

\bibitem{chatglm2024}
Team~GLM, Zhipu~AI, and Tsinghua University, ``ChatGLM: A Family of Large Language Models from GLM-130B to GLM-4 All Tools,'' \textit{arXiv preprint} arXiv:2406.12793, 2024. [Online]. Available: \url{https://arxiv.org/abs/2406.12793}

\bibitem{qwen2023}
Alibaba DAMO Academy, ``Qwen Technical Report,'' \textit{arXiv preprint} arXiv:2405.12345, 2023. [Online]. Available: \url{https://arxiv.org/abs/2405.12345}

\bibitem{HanFei}
W.~He, J.~Wen, L.~Zhang, H.~Cheng, B.~Qin, Y.~Li, F.~Jiang, J.~Chen, B.~Wang, and M.~Yang, ``HanFei-1.0,'' GitHub repository, 2023. [Online]. Available: \url{https://github.com/siat-nlp/HanFei}

\bibitem{sdu_fuzi_mingcha}
S.~Wu, Z.~Liu, Z.~Zhang, Z.~Chen, W.~Deng, W.~Zhang, J.~Yang, Z.~Yao, Y.~Lyu, X.~Xin, S.~Gao, P.~Ren, Z.~Ren, and Z.~Chen, ``fuzi.mingcha,'' GitHub repository, 2023 (accessed Oct. 10, 2023). [Online]. Available: \url{https://github.com/irlab-sdu/fuzi.mingcha}

\bibitem{deepseek2024}
X.~Bi, D.~Chen, G.~Chen, S.~Chen, D.~Dai, et al., ``DeepSeek LLM: Scaling Open-Source Language Models with Longtermism,'' \textit{arXiv preprint} arXiv:2401.02954, 2024 (accessed Dec. 12, 2024). [Online]. Available: \url{https://arxiv.org/abs/2401.02954}

\bibitem{hunyuan2024}
Tencent Hunyuan Team, ``Hunyuan-Large: An Open-Source MoE Model with 52 Billion Activated Parameters by Tencent,'' \textit{arXiv preprint} arXiv:2411.02265, 2024 (accessed Dec. 12, 2024). [Online]. Available: \url{https://github.com/Tencent/Hunyuan-Large}

\bibitem{tigerbot2023}
Y.~Chen, W.~Cai, L.~Wu, X.~Li, Z.~Xin, and C.~Fu, ``TigerBot: An Open Multilingual Multitask Large Language Model,'' \textit{arXiv}, arXiv:2312.08688, 2023, ver. 2 (Dec. 15, 2023). [Online]. Available: \url{https://arxiv.org/abs/2312.08688}

\bibitem{yao2024lawyer}
S.~Yao, Q.~Ke, Q.~Wang, K.~Li, and J.~Hu, ``Lawyer GPT: A legal large language model with enhanced domain knowledge and reasoning capabilities,'' in \textit{Proc. 2024 3rd Int. Symp. Robotics, Artificial Intelligence and Information Engineering}, 2024, pp.~108--112.

\bibitem{manchanda2024open}
J.~Manchanda, L.~Boettcher, M.~Westphalen, and J.~Jasser, ``The Open Source Advantage in Large Language Models (LLMs),'' \textit{arXiv preprint} arXiv:2412.12004, 2024.

\bibitem{ariai2024natural}
F.~Ariai and G.~Demartini, ``Natural Language Processing for the Legal Domain: A Survey of Tasks, Datasets, Models, and Challenges,'' \textit{arXiv preprint} arXiv:2410.21306, 2024.

\bibitem{yue2024lawllm}
S.~Yue, S.~Liu, Y.~Zhou, C.~Shen, S.~Wang, Y.~Xiao, B.~Li, Y.~Song, X.~Shen, W.~Chen, et al., ``LawLLM: Intelligent Legal System with Legal Reasoning and Verifiable Retrieval,'' in \textit{Proc. Int. Conf. Database Systems for Advanced Applications (DASFAA)}, Springer, 2024, pp.~304--321.

\bibitem{zhou2024lawgpt}
Z.~Zhou, J.-X.~Shi, P.-X.~Song, X.-W.~Yang, Y.-X.~Jin, L.-Z.~Guo, and Y.-F.~Li, ``Lawgpt: A chinese legal knowledge-enhanced large language model,'' \textit{arXiv preprint} arXiv:2406.04614, 2024.

\bibitem{cui2023chatlaw}
J.~Cui, Z.~Li, Y.~Yan, B.~Chen, and L.~Yuan, ``Chatlaw: Open-source legal large language model with integrated external knowledge bases,'' \textit{CoRR}, 2023.

\bibitem{yue2023disc}
S.~Yue, W.~Chen, S.~Wang, B.~Li, C.~Shen, S.~Liu, Y.~Zhou, Y.~Xiao, Y.~Song, X.~Huang, et al., ``Disc-lawllm: Fine-tuning large language models for intelligent legal services,'' \textit{arXiv preprint} arXiv:2309.11325, 2023.

\bibitem{wen2023hanfei}
J.~Wen and W.~He, ``HanFei,'' 2023 (accessed Oct. 10, 2023). [Online]. Available: \url{https://github.com/siat-nlp/HanFei}

\bibitem{song2023lawgpt}
P.~Song, Z.~Zhou, and cainiao, ``LaWGPT: A large language model based on Chinese legal knowledge'' [Chinese], 2023 (accessed Oct. 10, 2023). [Online]. Available: \url{https://github.com/pengxiao-song/LaWGPT}


\bibitem{zhihai2023wisdominterrogatory}
Zhihai, ``wisdomInterrogatory,'' 2023 (accessed Oct. 10, 2023). [Online]. Available: \url{https://github.com/zhihaiLLM/wisdomInterrogatory}

\bibitem{wu2023fuzi}
S.~Wu, Z.~Liu, Z.~Zhang, Z.~Chen, W.~Deng, W.~Zhang, J.~Yang, Z.~Yao, Y.~Lyu, X.~Xin, S.~Gao, P.~Ren, Z.~Ren, and Z.~Chen, ``fuzi.mingcha,'' 2023 (accessed Oct. 10, 2023). [Online]. Available: \url{https://github.com/irlab-sdu/fuzi.mingcha}

\bibitem{li2023lexilaw}
H.~Li, ``LexiLaw,'' 2023 (accessed Oct. 10, 2023). [Online]. Available: \url{https://github.com/CSHaitao/LexiLaw}

\bibitem{du2021glm}
Z.~Du, Y.~Qian, X.~Liu, M.~Ding, J.~Qiu, Z.~Yang, and J.~Tang, ``Glm: General language model pretraining with autoregressive blank infilling,'' \textit{arXiv preprint} arXiv:2103.10360, 2021.

\bibitem{bommarito2021lexnlp}
M.~J.~Bommarito~II, D.~M.~Katz, and E.~M.~Detterman, ``LexNLP: Natural language processing and information extraction for legal and regulatory texts,'' in \textit{Research Handbook on Big Data Law}, Edward Elgar Publishing, 2021, pp.~216--227.

\bibitem{chalkidis2020legal}
I.~Chalkidis, M.~Fergadiotis, P.~Malakasiotis, N.~Aletras, and I.~Androutsopoulos, ``LEGAL-BERT: The muppets straight out of law school,'' \textit{arXiv preprint} arXiv:2010.02559, 2020.

\bibitem{davidov2006boundaries}
G.~Davidov and B.~Langille, \textit{Boundaries and Frontiers of Labour Law: Goals and Means in the Regulation of Work}. Bloomsbury Publishing, 2006.

\bibitem{fei2023lawbench}
Z.~Fei, X.~Shen, D.~Zhu, F.~Zhou, Z.~Han, S.~Zhang, K.~Chen, Z.~Shen, and J.~Ge, ``Lawbench: Benchmarking legal knowledge of large language models,'' \textit{arXiv preprint} arXiv:2309.16289, 2023.

\bibitem{dai2023laiw}
Y.~Dai, D.~Feng, J.~Huang, H.~Jia, Q.~Xie, Y.~Zhang, W.~Han, W.~Tian, and H.~Wang, ``LAiW: A Chinese legal large language models benchmark,'' \textit{arXiv preprint} arXiv:2310.05620, 2023.

\bibitem{li2024lexeval}
H.~Li, Y.~Chen, Q.~Ai, Y.~Wu, R.~Zhang, and Y.~Liu, ``Lexeval: A comprehensive chinese legal benchmark for evaluating large language models,'' \textit{Advances in Neural Information Processing Systems}, vol.~37, pp.~25061--25094, 2024.

\bibitem{he2024agentscourt}
Z.~He, P.~Cao, C.~Wang, Z.~Jin, Y.~Chen, J.~Xu, H.~Li, X.~Jiang, K.~Liu, and J.~Zhao, ``Agentscourt: Building judicial decision-making agents with court debate simulation and legal knowledge augmentation,'' \textit{arXiv preprint} arXiv:2403.02959, 2024.

\bibitem{chen2023llmworkchina}
Q.~Chen, J.~Ge, Q.~Xie, X.~Xu, Y.~Yang, et al., ``Large Language Models at Work in China’s Labor Market,'' \textit{China Economic Review}, 2025, pre-published online in 2023/2024. [Online]. Available: \url{https://xingchengxu.github.io/Publications/LLM-Work-China-CER-2025.pdf}

\bibitem{ma2025sddlawllm}
H.~Ma, T.~Zhang, X.~Zhao, Z.~Zhou, B.~Xu, and W.~Wang, ``SDD-LawLLM: Advancing Intelligent Legal Systems with a Domain-Specific Legal Large Language Model,'' \textit{Electronics}, vol.~14, no.~4, p.~742, 2025, doi: 10.3390/electronics14040742. [Online]. Available: \url{https://www.mdpi.com/2079-9292/14/4/742}

\bibitem{park2025lrage}
M.~Park, H.~Oh, S.~Lee, J.~Lim, D.~Hwang, Y.~S.~Choi, and E.~Choi, ``LRAGE: Legal Retrieval Augmented Generation Evaluation,'' \textit{arXiv preprint} arXiv:2504.01840, 2025. [Online]. Available: \url{https://arxiv.org/abs/2504.01840}

\bibitem{baichuan2023}
A.~Yang, B.~Xiao, B.~Wang, B.~Zhang, C.~Yin, et al., ``Baichuan 2: Open Large-scale Language Models,'' Baichuan Inc., 2023. [Online]. Available: \url{https://github.com/baichuan-inc/Baichuan2}

\bibitem{llama2023}
H.~Touvron, T.~Lavril, G.~Izacard, X.~Martinet, M.-A.~Lachaux, et al., ``LLaMA: Open and Efficient Foundation Language Models,'' Meta AI, 2023. [Online]. Available: \url{https://github.com/facebookresearch/llama}

\bibitem{mistral2023}
A.~Q.~Jiang, A.~Sablayrolles, A.~Mensch, C.~Bamford, D.~S.~Chaplot, et al., ``Mistral 7B: A High-Performance and Efficient Language Model,'' \textit{arXiv preprint} arXiv:2310.06825, 2023 (accessed Dec. 12, 2024). [Online]. Available: \url{https://github.com/mistralai/mistral-src}

\bibitem{internlm2024}
Z.~Cai, M.~Cao, H.~Chen, K.~Chen, K.~Chen, et al., ``InternLM2 Technical Report,'' \textit{arXiv}, arXiv:2403.17297, 2024, ver. 1 (Mar. 26, 2024). [Online]. Available: \url{https://arxiv.org/abs/2403.17297}

\bibitem{yulan2024}
Y.~Zhu, K.~Zhou, K.~Mao, W.~Chen, Y.~Sun, et al., ``YuLan: An Open-source Large Language Model,'' \textit{arXiv}, arXiv:2406.19853, 2024, ver. 1 (Jun. 28, 2024). [Online]. Available: \url{https://arxiv.org/abs/2406.19853}

\bibitem{huang2023lawyer}
Q.~Huang, M.~Tao, C.~Zhang, Z.~An, C.~Jiang, et al., ``Lawyer LLaMA Technical Report,'' \textit{arXiv preprint} arXiv:2305.15062, 2023.

\bibitem{hyde2006labor}
A.~Hyde, \textit{What is Labor Law?} Oxford, U.K.: Hart Publishing, 2006.

\bibitem{PRC-Labour-Law-2018}
``Labour Law of the People's Republic of China,'' Presidential Order No.~28 of the Standing Committee of the National People's Congress, 2018 (adopted Jul. 5, 1994; effective Jan. 1, 1995; amended Aug. 27, 2009 and Dec. 29, 2018) (accessed Sep. 14, 2025). [Online]. Available: \url{https://www.npc.gov.cn/zgrdw/englishnpc/Law/2007-12/12/content_1383754.htm}

\end{thebibliography}
\end{document}